\documentclass[letterpaper,10pt,conference]{ieeeconf}
\IEEEoverridecommandlockouts
\usepackage{xcolor}
\makeatletter
\let\NAT@parse\undefined
\makeatother
\usepackage[numbers,sort&compress]{natbib}
\usepackage{algorithm}
\usepackage{algorithmic}
\usepackage{microtype}
\usepackage{graphicx}
\usepackage{subcaption}
\usepackage{booktabs} %
\usepackage{placeins}
\usepackage{pgfplots}
\pgfplotsset{compat=1.18}
\usepackage{xspace}
\usepackage{xfp}
\usepackage[colorlinks=true,linkcolor={blue!60!black},citecolor=black,urlcolor=black]{hyperref}
\usepackage[all]{hypcap}

\usepackage{bbm}

\title{\LARGE\bfseries Video Replanning via Latent Embedding Refinement\\and Rejection-Based Sampling}

\usepackage{amsmath}
\usepackage{amssymb}
\usepackage{mathtools}

\usepackage{amsthm}
\usepackage{tikz}
\pgfplotsset{compat=1.18}

\newcommand{\iseFullandAbbr}{RELIC: REplanning with Latent embedding refInement and Candidate rejection}
\newcommand{\ise}{RELIC}
\newcommand{\replanentry}[2]{$#1${\scriptsize\color{black!55}$\pm$#2}} %

\newcommand{\opendoor}{\textsc{OpenDoor}\xspace}
\newcommand{\slideblock}{\textsc{SlideBlock}\xspace}
\newcommand{\balance}{\textsc{Balance}\xspace}
\newcommand{\pinball}{\textsc{Pinball}\xspace}
\newcommand{\pushbar}{\textsc{PushBar}\xspace}
\newcommand{\pickbar}{\textsc{PickBar}\xspace}
\newcommand{\slidebrick}{\textsc{SlideBrick}\xspace}
\renewcommand{\openbox}{\textsc{OpenBox}\xspace}
\newcommand{\turnfaucet}{\textsc{TurnFaucet}\xspace}
  
\newcommand{\ourwebsite}{https://RELIC-video.github.io}

\newcommand{\ICLRRevision}[1]{\textcolor{black}{#1}}

\usepackage{amsmath,amsfonts,bm}

\newcommand{\ie}{\textit{i}.\textit{e}.,\ }
\newcommand{\eg}{\textit{e}.\textit{g}.,\ }

\def\eqref#1{equation~\ref{#1}}

\def\1{\bm{1}}

\DeclareMathAlphabet{\mathsfit}{\encodingdefault}{\sfdefault}{m}{sl}
\SetMathAlphabet{\mathsfit}{bold}{\encodingdefault}{\sfdefault}{bx}{n}

\usepackage{url}
\usepackage[capitalize,noabbrev,nameinlink]{cleveref}
\usepackage{wrapfig}
\crefname{section}{Section}{Sections}
\Crefname{section}{Section}{Sections}
\Crefname{table}{Table}{Tables}
\crefname{table}{Table}{Tables}
\Crefname{figure}{Figure}{Figures}
\crefname{figure}{Figure}{Figures}
\crefname{subfigure}{Figure}{Figures}
\Crefname{subfigure}{Figure}{Figures}

\author{Po-Chen Ko$^{1}$, Yu Chan$^{1}$, Yu-Hsiang Fu$^{1}$, Hsien-Jeng Yeh$^{1}$, Chu-Rong Chen$^{1}$,\\ Dong-Chen Tsai$^{2}$, Wei-Chiu Ma$^{3}$, Yilun Du$^{4}$, Jiayuan Mao$^{5}$, and Shao-Hua Sun$^{1}$%
\thanks{$^{1}$National Taiwan University. $^{2}$Industrial Technology Research Institute, Taiwan. $^{3}$Cornell University. $^{4}$Harvard University. $^{5}$MIT CSAIL.}%
\thanks{Preprint. An earlier version of this manuscript appeared under the title ``Implicit State Estimation via Video Replanning''; the method was then called ISE.}%
}

\begin{document}
\bstctlcite{IEEEexample:BSTcontrol} 

\maketitle

\begin{abstract}
\label{sec:abstract}
Video planning has emerged as a flexible framework for robot manipulation, in which a generative model predicts a video of task completion, and a downstream module translates the predicted frames into actions. 
However, existing methods typically ignore information from past interactions, limiting their ability to adapt to latent physical properties that can only be revealed through trial and error, such as whether a door should be pushed or pulled, or how friction affects object dynamics. 
When a plan fails, these methods usually replan from scratch without leveraging the information revealed by the failure.
To address this limitation, we introduce RELIC, REplanning with Latent embedding refInement and Candidate rejection, a video planning framework that adapts to hidden physical properties from test-time failures. 
RELIC optimizes a latent embedding that captures the environment's hidden physical properties from interaction videos and introduces a rejection-based sampling mechanism that filters out hypotheses inconsistent with prior failures. 
Across eight tasks in two simulation suites, RELIC consistently reduces the number of replanning steps required for success, and linear probes show that its embedding captures the hidden parameters from the interaction itself rather than from scene appearance.
Across four challenging real-world robotic manipulation tasks involving hidden interaction modes, \eg friction, center of mass, and object mass, RELIC raises the one-shot replanning success rate of a video planning baseline from $30.0\%$ to $63.8\%$ after a single physical interaction.
\end{abstract}

\section{Introduction}
\label{sec:intro}

Video has emerged as a powerful representation for robot learning and planning, capturing both perception and action across tasks and environments~\cite{okami2024}. In video planning, a system imagines task execution conditioned on a goal and current observation, then converts the predicted frames into robot actions~\cite{chang2020procedure,du2024learning}. Prior work makes these plans executable with predicted frames, correspondences, or point tracks that guide downstream policies~\cite{ko2024learning,wen2023any,bharadhwaj2025track2act}. Yet accurate plan following is insufficient when the plan rests on a wrong assumption.

Physical properties such as friction, mass, or a door's opening direction may be hidden from the current observation. A planner may therefore prescribe pushing a door that can only be pulled; better execution of that motion cannot correct the mistaken assumption. The failed interaction, however, provides evidence that can guide the next plan. VLM-guided planning~\cite{zhao2024vlmpc} and history-aware verification~\cite{li2025learn} support candidate evaluation and selection. We address a complementary challenge: using interaction outcomes to revise the latent hypothesis that conditions video-plan generation.

One approach is to train explicit belief models to infer hidden parameters from simulation~\cite{memmel2024asid,qi2023hand}, but this requires specifying which parameters to estimate. Our goal is to adapt video planning from past interactions without a predefined parameterization of the hidden physical properties. We propose \iseFullandAbbr{}, illustrated in \Cref{fig:video_replanning}. \ise{} retrieves and refines a latent embedding that implicitly captures these properties, then conditions video-plan generation on this embedding. Failed interactions provide evidence for refinement, and a rejection mechanism discards candidates resembling past failures.

\definecolor{USCRed}{RGB}{153, 27, 30}
\definecolor{BerkeleyBlue}{RGB}{0, 38, 118}
\begin{figure}[t]
  \centering
  \includegraphics[width=\linewidth]{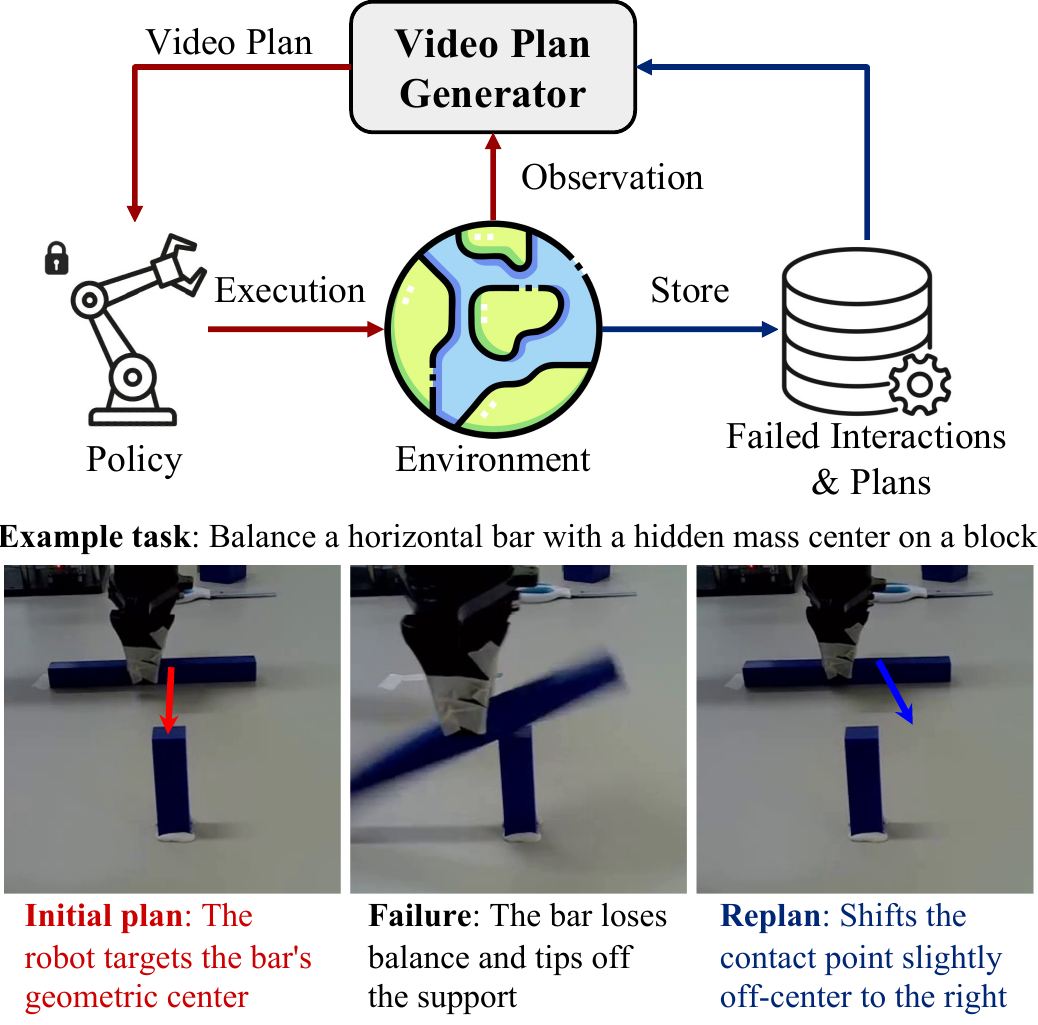}
  \caption{\textbf{Video replanning from test-time failures.}
  Top: existing video planning methods condition only on the first frame (\textcolor{USCRed}{red}). RELIC also stores failed interactions and plans (\textcolor{BerkeleyBlue}{blue}) and uses them to generate the next plan. Bottom: a real-robot example. The bar's center of mass is hidden, so the first plan fails; the failure reveals it, and RELIC replans off-center.}
  \label{fig:video_replanning}
  \vspace{-6pt}
\end{figure}

\begin{figure*}[!t]
    \centering
    \includegraphics[clip, trim=0.0cm 9.2cm 69.2cm 0.0cm, width=\linewidth]{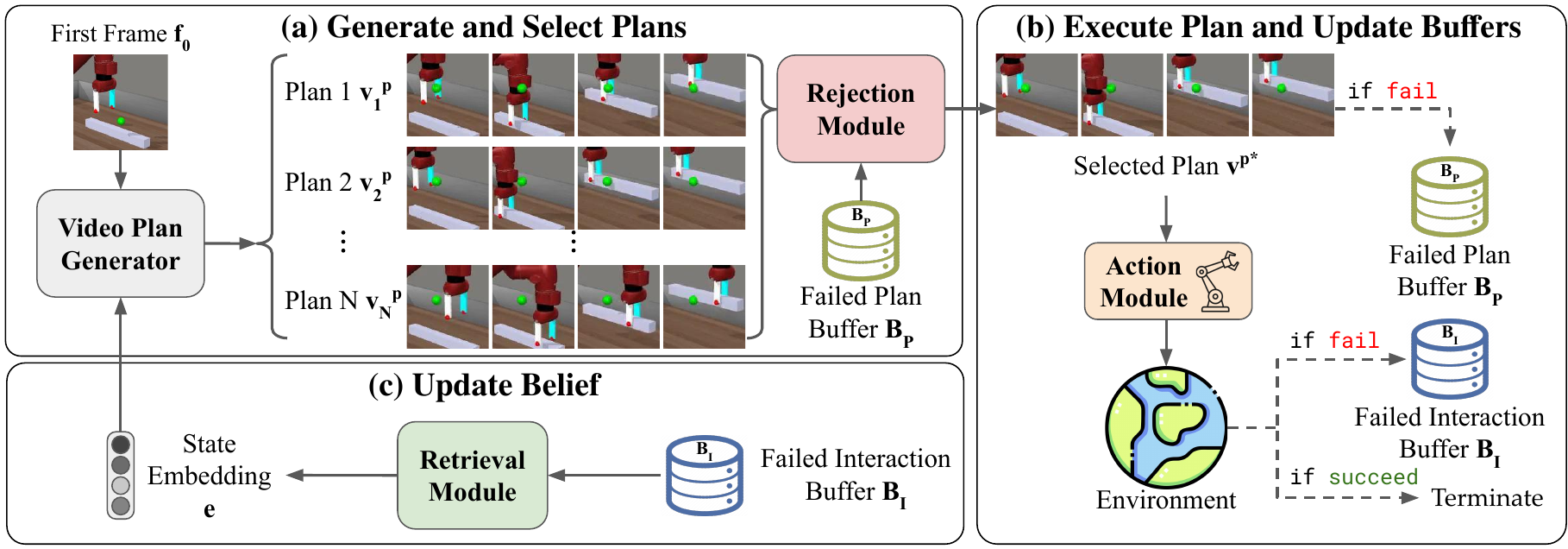}
    \caption{\textbf{Framework overview.} \textbf{(a) Generate and select plans.} Conditioned on the first frame $f_0$ and embedding $e$, the Video Plan Generator proposes candidate plans; the Rejection Module selects the one least similar to the failed plans in $B_P$. \textbf{(b) Execute and update buffers.} The Action Module executes the selected plan $v^{p*}$. On failure, the plan and observed interaction are added to $B_P$ and $B_I$, respectively. \textbf{(c) Update belief.} Retrieval and refinement update $e$ using $B_I$ before the next planning round.}
    \label{fig:framework}
\end{figure*}

Our contributions are:
\begin{itemize}\setlength{\itemsep}{0pt}\setlength{\parskip}{0pt}\setlength{\topsep}{0pt}
    \item \textbf{RELIC}, a video planning framework that refines a latent embedding from failed interactions and rejects candidate plans resembling past failures; linear probes show that this embedding captures the hidden parameter from the interaction rather than from scene appearance.
    \item \textbf{Two system-identification suites} of eight tasks on Meta-World~\cite{Yu2019MetaWorldAB} and ManiSkill3~\cite{taomaniskill3}, with hidden mass, center of mass, friction, elasticity, or interaction mode, where RELIC requires the fewest replans.
    \item \textbf{Real-world validation} on four tasks and two robot arms, raising one-shot replanning success from $30.0\%$ to $63.8\%$.
\end{itemize}

\section{Related Work}
\paragraph{Video planning}
Videos have become a popular medium for planning in the physical world~\cite{yang2024position}. Prior approaches use video to train dynamics models~\cite{yang2024learning} or reward models~\cite{rwm1}, directly predict video sequences~\cite{du2024learning,ko2024learning,black2024zeroshot}, or predict derived representations such as 3D correspondences or point tracks to guide a downstream policy~\cite{wen2023any,bharadhwaj2025track2act}. Most, however, ignore environmental uncertainty, limiting their ability to adapt and replan.

\paragraph{System identification}
System identification estimates system dynamics and parameters from experimental data, with a long history of active, information-maximizing strategies~\cite{ljung1998system} and recent applications to real-world physical-parameter estimation~\cite{physparam1,physparam2,memmel2024asid}. Verifier-based frameworks~\cite{li2025learn} that reject infeasible action proposals from historical input have also enabled online adaptation. We bridge classical system identification and modern video planning by implicitly inferring physical parameters from past video data.

\paragraph{Few-shot adaptation and latent context inference}
A common abstraction for few-shot adaptation models task variation through latent context or hidden parameters. Hidden-parameter MDPs~\cite{killian2017hipmdp} formalize environment variation via unobserved parameters inferred from interaction, a view extended in meta-RL through latent-context representations~\cite{rakelly2019efficient}. In robotics, related methods address dynamics shifts and embodiment variation~\cite{kumar2021rma, gao2025adaworld}, and retrieval-based approaches reuse past interactions via sub-trajectory selection~\cite{memmel2025strap}. We likewise treat system variation as latent, but rather than maintaining beliefs to tune policies, we optimize the model's internal representation directly for test-time planning.
\section{Approach}
\label{sec:method}

We consider settings where task-relevant system parameters $\theta$, such as an object's mass, surface friction, or whether a door pushes or pulls, are unknown at test time. Unlike prior work on explicit state estimation, we do not require prior knowledge of what $\theta$ represents; we only assume it can be inferred from interaction data. For each environment, we assume access to an experience dataset $\mathcal{D} = \{(v_i, o_i, s_i)\}$, where $v_i$ is an interaction video, $o_i$ is a \emph{configuration ID} grouping trials known to share underlying dynamics, and $s_i \in \{0,1\}$ indicates whether the interaction succeeded. Configuration IDs only group trials that share $\theta$; they are never used as labels for $\theta$. $\mathcal{D}$ contains both successful and failed interactions. Our goal is to generate video plans that, when executed by a fixed Action Module, yield successful action sequences.

As shown in \Cref{fig:framework}, given the current scene and past interactions, it retrieves and refines a latent embedding that implicitly captures the structure of the hidden parameters, then generates a video plan conditioned on this embedding. We maintain two test-time buffers: $B_I$ for failed interactions and $B_P$ for failed plans. At each round, the framework updates the latent embedding using $B_I$, as described in \Cref{sec:method:planner}, generates $N$ candidate plans, and selects the one most different from $B_P$ via the rejection strategy of \Cref{sec:method:rejection_module}. The Action Module of \Cref{sec:method:action_module} converts the chosen plan into executable actions; we instantiate it with the training-free point-tracking method from AVDC~\cite{ko2024learning}. \Cref{sec:apdx:method} details each component.

\paragraph{Plans vs.\ interactions}
Although both are represented as videos, \textbf{plans} and \textbf{interactions} play asymmetric roles. A \textbf{plan} is generated by the planner trained only on successful executions; when its hypothesis about the environment is wrong, it may produce physically implausible rollouts, such as a video of pushing a door that should be pulled. An \textbf{interaction} is a real execution and faithfully reflects the true underlying dynamics, though it may be suboptimal. This asymmetry motivates maintaining the two buffers separately: failed plans expose flawed hypotheses, while failed interactions provide a ground-truth signal about the environment.

\begin{figure}[!t]
    \centering
    \includegraphics[clip, trim=0.0cm 10.8cm 82.8cm 0.0cm, width=\linewidth]{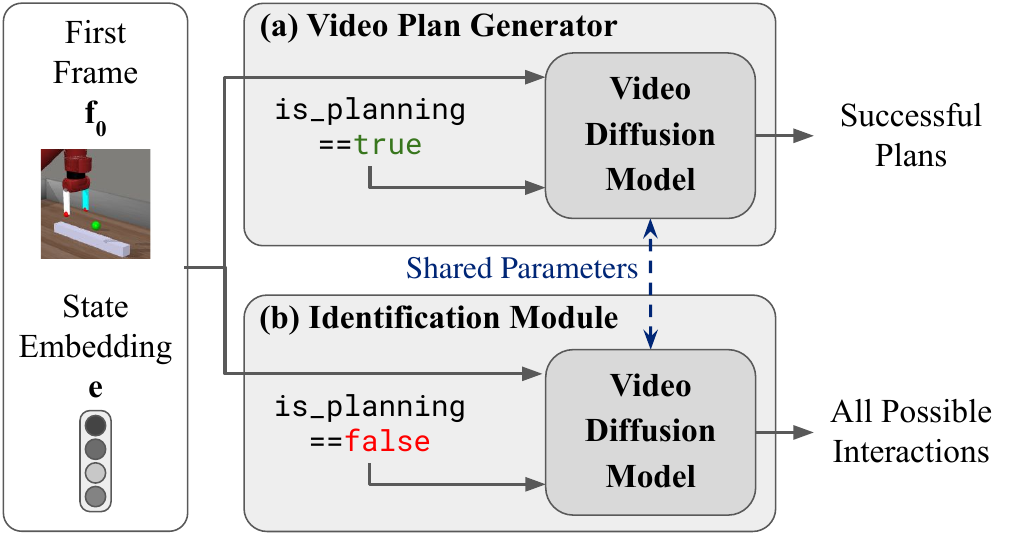}
    \caption{\textbf{Video Plan Generator and Identification Module.} The two modules share parameters and are selected by the binary trigger \texttt{is\_planning}. The planner learns from successful videos in $\mathcal{D}$; the identification module learns from all interactions, including failures, to support state-embedding refinement.}
    \label{fig:model}
\end{figure}

\subsection{Video-based Planning with State Embeddings}
\label{sec:method:planner}
\vspace{-5pt}
To plan effectively under unknown dynamics, our system learns a latent representation that implicitly captures hidden parameters such as mass or friction, through \emph{state embeddings}.

\paragraph{State embeddings}
We represent the underlying physical properties of an interaction by encoding its video frame-by-frame with a pretrained image encoder $E$ and concatenating the per-frame features, giving a per-video \emph{state embedding} $e_i = E(v_i)$ for every trial in $\mathcal{D}$, including both successful and failed ones. We use DINOv2~\citep{oquab2024dinov} as the default encoder and also experiment with R3M~\citep{nair2022rm} and CLIP~\citep{radford2021learning}. For each configuration ID, we additionally select one successful trial and designate its state embedding as that configuration's \emph{canonical embedding} $e^o$, which serves as the conditioning signal for the planner. The two roles are complementary: state embeddings of failed trials remain useful for retrieval, as we describe next, while canonical embeddings, drawn from successful trials, align with the planner, which is itself trained only on successful executions.

\paragraph{Retrieval and refinement}
At inference time, the planner requires a canonical embedding aligned with the current interaction. After a failed attempt, we encode the resulting interaction with $E$ and compute similarity scores against the stored per-video embeddings $\{e_i\}$ (which include embeddings of failed trials). We apply a softmax over these scores to obtain a distribution over $\{e_i\}$, sample a per-video embedding, and return its associated canonical embedding $e^{o_i}$ for use by the planner. We retrieve using the most recent failed interaction, with cosine distance on DINOv2 features and softmax temperature $\tau=0.3$. Aggregating evidence over earlier interactions can also help but is less consistent, as analyzed in \Cref{sec:apdx:multiple_interaction_retrieval}.

Retrieval returns the closest stored configuration. To adapt it to the specific interaction, including $\theta$ values absent from $\mathcal{D}$, we add an optimization-based refinement step, shown in \Cref{fig:model}: a generative \emph{Identification Module} that shares parameters with the Video Plan Generator is trained to produce both successful and unsuccessful rollouts. At inference, we freeze the module's weights and optimize the embedding using a denoising loss: the MSE between the observed interaction video and the module's clean-video prediction from a noisy version of that interaction, conditioned on the embedding. Only the embedding is updated, using 100 Adam steps with learning rate $10^{-3}$. The refinement can also operate without retrieval by initializing from a multivariate Gaussian prior fit to the stored canonical embeddings; we report this variant as ``RELIC (Refine FS)'' in experiments.

\paragraph{Video plan generation}
Once a canonical embedding is obtained, the \emph{Video Plan Generator} predicts a sequence of plausible futures conditioned on the current observation and the embedding. We adopt the diffusion-based planner of AVDC~\cite{ko2024learning} and concatenate its language conditioning with our canonical embeddings, projected into the CLIP-Text token space and appended to the 3D U-Net backbone. The generator predicts seven future frames at $32\times32$ resolution with 25 DDIM sampling steps; a residual super-resolution model upsamples the plan to $128\times128$. Training and sampling details are in \Cref{sec:apdx:method:planner:video_generator}.

\begin{figure}[!t]
    \centering
    \includegraphics[clip, trim=5bp 8bp 1820bp 5bp, width=\linewidth]{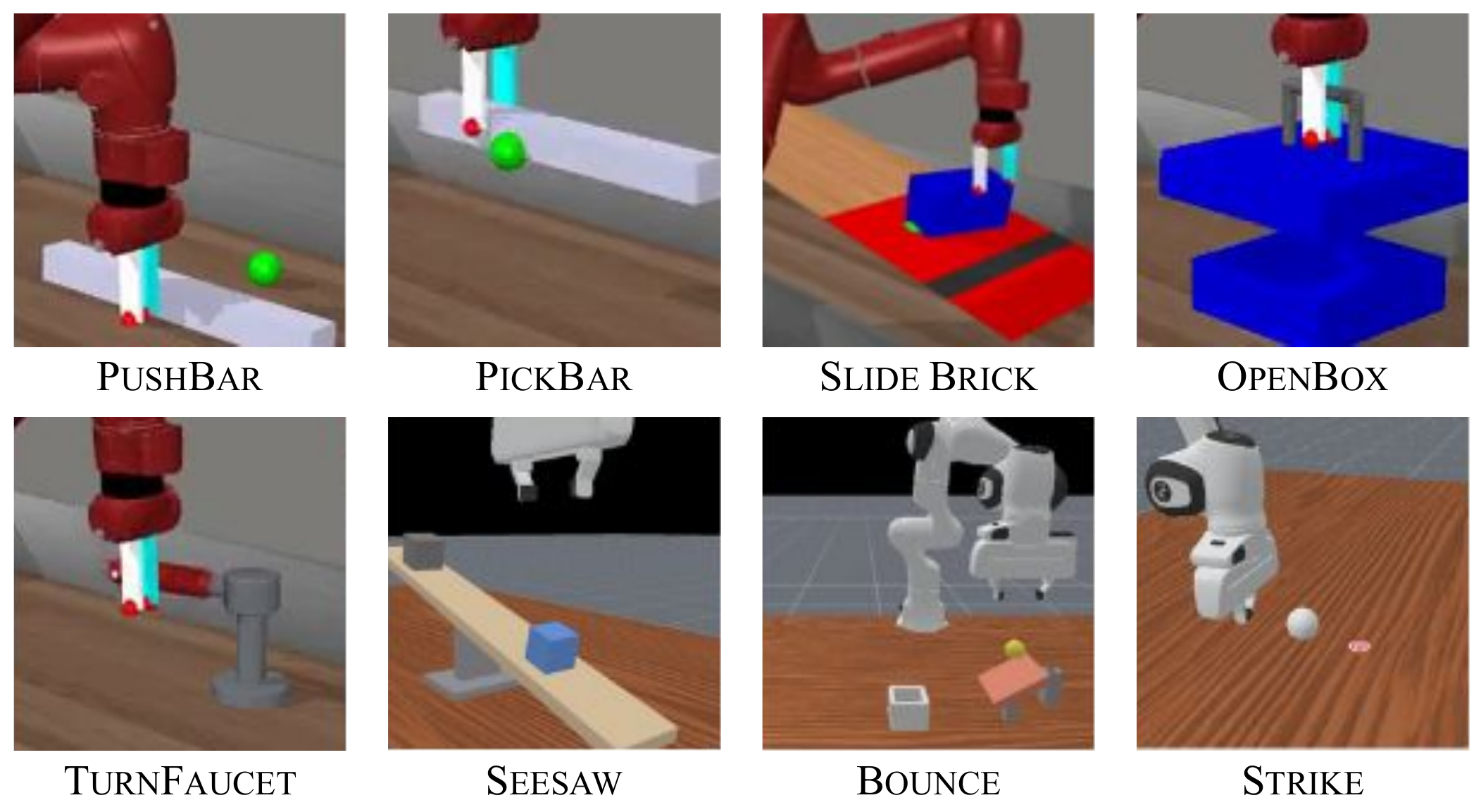}
    \caption{\textbf{Simulation tasks.} Five tasks from our Meta-World suite are followed by three from our ManiSkill3 suite. Each panel shows a representative task frame; the tasks require adaptation to hidden physical properties.}
    \label{fig:sim-suite-overview}
\end{figure}

\begin{table*}[!t]
    \caption{\textbf{Replanning performance across our two simulation task suites} ($\downarrow$).
    Per-task columns report raw replanning counts as mean $\pm$ SEM over 400 trials; CQL also averages three policies.
    \textit{Overall} reports the mean ratio of each method's replanning count to that of RELIC across the eight tasks, with equal weight for each task; its SEM is propagated from the per-task SEMs by the delta method, assuming independence across methods.}
    \label{tab:mw_bench_envs}
    \centering
    \footnotesize
    \setlength{\tabcolsep}{2.5pt}
    \renewcommand{\arraystretch}{1.18}
    \begin{tabular*}{\textwidth}{@{\extracolsep{\fill}}l*{9}{c}@{}}
    \toprule
    & \multicolumn{5}{c}{\textbf{Meta-World}} & \multicolumn{3}{c}{\textbf{ManiSkill3}} & \\
    \cmidrule(lr){2-6}\cmidrule(lr){7-9}
    \textbf{Method} & \pushbar & \pickbar & \slidebrick & \openbox & \turnfaucet & \textsc{Seesaw} & \textsc{Bounce} & \textsc{Strike} & \shortstack{\textbf{Overall}\\[-2pt]{\scriptsize (all 8, norm.)}} \\
    \midrule
    CQL & \replanentry{9.96}{0.20} & \replanentry{9.22}{0.19} & \replanentry{8.94}{0.18} & \replanentry{3.32}{0.18} & \replanentry{5.63}{0.21} & \replanentry{5.91}{0.18} & \replanentry{6.03}{0.68} & \replanentry{4.96}{0.37} & \replanentry{2.12}{0.08} \\
    BC & \replanentry{10.26}{0.30} & \replanentry{8.79}{0.32} & \replanentry{9.24}{0.33} & \replanentry{1.59}{0.19} & \replanentry{3.54}{0.27} & \replanentry{5.17}{0.09} & \replanentry{7.48}{0.71} & \replanentry{4.02}{0.20} & \replanentry{1.74}{0.06} \\
    AVDC & \replanentry{6.83}{0.27} & \replanentry{4.41}{0.22} & \replanentry{7.36}{0.27} & \replanentry{1.82}{0.16} & \replanentry{1.67}{0.16} & \replanentry{5.62}{0.47} & \replanentry{5.80}{0.48} & \replanentry{4.13}{0.76} & \replanentry{1.29}{0.06} \\
    VLM verifier & \replanentry{4.73}{0.40} & \replanentry{4.26}{0.38} & \replanentry{7.35}{0.45} & \replanentry{1.56}{0.36} & \replanentry{1.41}{0.16} & \replanentry{5.72}{0.48} & \replanentry{5.64}{0.43} & \replanentry{4.12}{0.41} & \replanentry{1.17}{0.06} \\
    \midrule
    RELIC (Refine FS) & \replanentry{4.66}{0.23} & \replanentry{3.87}{0.21} & \replanentry{6.72}{0.26} & \replanentry{1.73}{0.16} & \replanentry{1.54}{0.15} & \replanentry{5.31}{0.37} & \replanentry{5.32}{0.38} & \replanentry{2.96}{0.26} & \replanentry{1.10}{0.04} \\
    \textbf{RELIC (Ours)} & \replanentry{\mathbf{4.26}}{0.20} & \replanentry{\mathbf{3.84}}{0.18} & \replanentry{\mathbf{6.69}}{0.26} & \replanentry{\mathbf{1.25}}{0.10} & \replanentry{\mathbf{1.39}}{0.14} & \replanentry{\mathbf{4.88}}{0.37} & \replanentry{\mathbf{5.12}}{0.40} & \replanentry{\mathbf{2.72}}{0.29} & \textbf{1.00} \\
    \bottomrule
    \end{tabular*}
\end{table*}

\begin{figure*}[t]
    \centering
    \includegraphics[clip, trim=0bp 204bp 1807bp 0bp, width=0.95\linewidth]{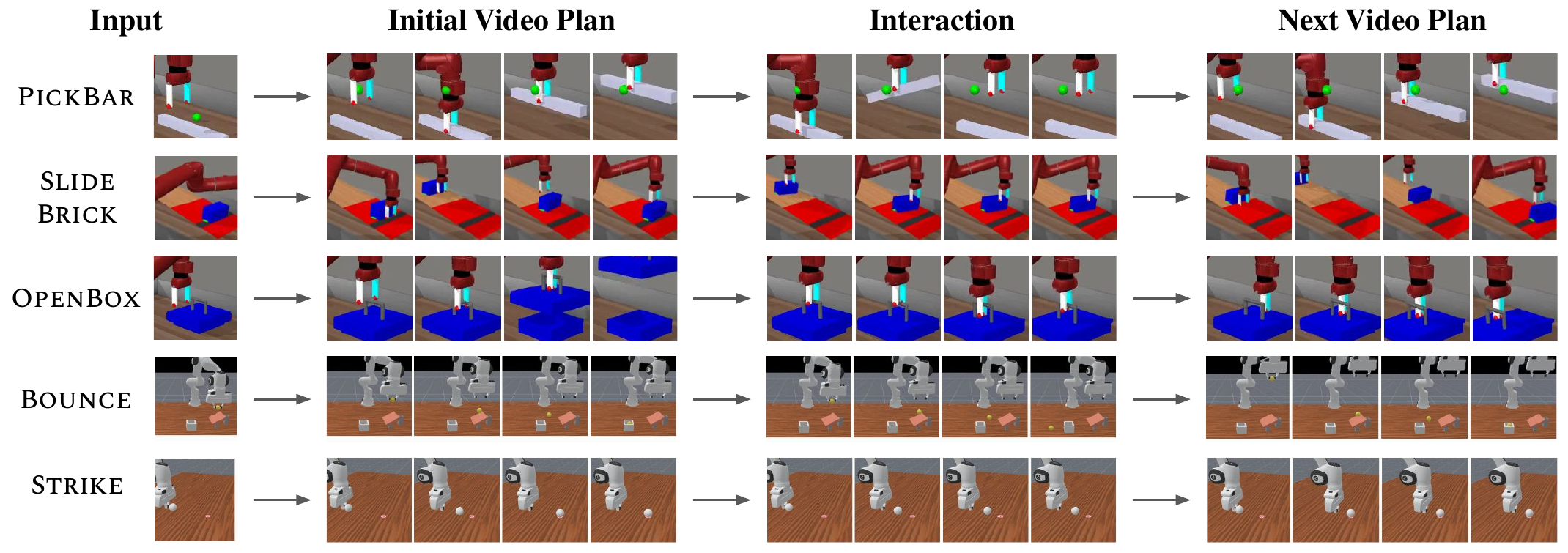}
    \caption{\textbf{Qualitative adaptation examples on simulation tasks.} Rows show \pickbar, \slidebrick, and \openbox from our Meta-World suite, followed by \textsc{Bounce} and \textsc{Strike} from our ManiSkill3 suite. Columns show the input, initial plan, interaction, and revised plan. ManiSkill3 plan panels are simulator renderings under the plan's hypothesized dynamics; interaction panels use the true dynamics.}
    \label{fig:qual}
\end{figure*}

\subsection{Rejection-based Sampling}
\label{sec:method:rejection_module}
A single embedding commits the planner to one hypothesis about $\theta$. Considering several hypotheses at once avoids this commitment and calls for a mechanism that selects among the resulting plans. Video planning naturally supplies the material for one: past plans are videos, and failed ones can be compared directly against new candidates. We add a \emph{Rejection Module} that discards candidates resembling plans that have already failed. At each planning round, the Video Plan Generator produces $N$ candidate plans $\{v^p_1, \ldots, v^p_N\}$ by conditioning on $N$ independently sampled embeddings, so each candidate corresponds to a different hypothesis. Using the failed plan buffer $B_P$, we select the candidate furthest from any prior failure:
\[
v^{p*} = \arg\max_{v^p_i} \min_{F \in B_P} K_{rej}(v^p_i, F),
\]
where $K_{rej}$ is a distance function; we use raw-pixel L2 distance by default and compare alternatives in \Cref{fig:ablation-3}. This steers the planner toward strategies that differ from past failures rather than repeating them.

\subsection{Action Module}
\label{sec:method:action_module}

We follow the dense object- and wrist-tracking approach from AVDC~\cite{ko2024learning}: the predicted video frames are used to track either the object of interest or, in tasks such as bar picking, where object tracking is insufficient, the robot's wrist. The resulting trajectory is converted into control commands via simple heuristic policies such as grasp strategies or trajectory-following controllers. This training-free approach enables direct execution of video plans without an additional learned policy.
\section{Experiments}

\subsection{Experimental Setup}
\label{sec:exp:env}

\begin{table}[t]
    \caption{\textbf{Offline experience dataset $\mathcal{D}$ per suite.} Total number of interaction videos and the number of successful ones; a video is successful when the guessed value equals the true value. Each video is one scripted execution in which the policy assumes a guessed parameter value while the environment uses a true hidden value.}
    \label{tab:dataset}
    \centering
    \footnotesize
    \setlength{\tabcolsep}{3pt}
    \renewcommand{\arraystretch}{1.12}
    \begin{tabular*}{\columnwidth}{@{\extracolsep{\fill}}lccc@{}}
    \toprule
    \textbf{Suite} & \textbf{Tasks} & \textbf{Videos} & \textbf{Successful} \\
    \midrule
    Meta-World & 5 & 14,270 & 670 \\
    ManiSkill3 & 3 & 18,750 & 750 \\
    \bottomrule
    \end{tabular*}
\end{table}

\begin{table}[t]

  \caption{\textbf{Evaluation on replanned videos.} 
    Similarity of replanned videos to ground-truth videos from the scripted policy. Arrows indicate better performance; best values are bold.
  }
  \centering
  \label{tab:cv-metrics-left}
  \footnotesize
  \setlength{\tabcolsep}{3pt}
  \renewcommand{\arraystretch}{1.16}
  \begin{tabular*}{\linewidth}{@{\extracolsep{\fill}}lcccc@{}}
      \toprule 
      \textbf{Method} & \textbf{PSNR $\uparrow$} & \textbf{LPIPS $\downarrow$} & \textbf{CLIP $\uparrow$} & \textbf{DINO $\uparrow$} \\ 
      \midrule
      AVDC            & 19.426 & 0.271 & 0.870 & 0.705\\  
      RELIC (R3M)     & 19.632 & 0.261 & 0.875 & 0.705\\ 
      RELIC (CLIP)    & 19.584 & 0.262 & 0.871 & 0.707\\  
      \textbf{RELIC (DINOv2, Ours)} & \textbf{19.817} & \textbf{0.255} &\textbf{ 0.882} & \textbf{0.717}\\
      \bottomrule
  \end{tabular*}
\end{table}

\textbf{Simulation task suites.} We develop two system-identification task suites using Meta-World and ManiSkill3, respectively, to evaluate online adaptation under unknown system parameters $\theta$. Across both suites, agents must use interaction outcomes to adapt their plans to hidden physical properties. Both suites use the same replanning evaluation protocol. For data collection, hidden parameters are discretized; for example, the bar's center of mass takes 25 positions along its 0.4\,m length. At evaluation they are sampled continuously from the same intervals, so test values generally differ from those in $\mathcal{D}$.

Our Meta-World suite comprises \pushbar, \pickbar, \slidebrick, \openbox, and \turnfaucet; our ManiSkill3 suite comprises \textsc{Seesaw}, \textsc{Bounce}, and \textsc{Strike}. Together, the suites cover variations in mass in \textsc{Seesaw}, center of mass in \pushbar and \pickbar, friction in \slidebrick and \textsc{Strike}, elasticity in \textsc{Bounce}, and interaction mode in \openbox and \turnfaucet. \Cref{fig:sim-suite-overview} shows all eight environments, and \Cref{fig:qual} illustrates adaptation on five representative tasks across the two suites. Hidden-parameter ranges, train/test protocols, and success criteria for the Meta-World suite are given in \Cref{sec:apdx:benchmark}.

\begin{figure*}[!t]
    \centering
    \captionsetup[subfigure]{format=plain,justification=raggedright,singlelinecheck=false}

    \begin{subfigure}[t]{0.32\textwidth}
        \centering
        \makebox[\linewidth][c]{\raisebox{0mm}[40mm][0pt]{\includegraphics[height=40mm]{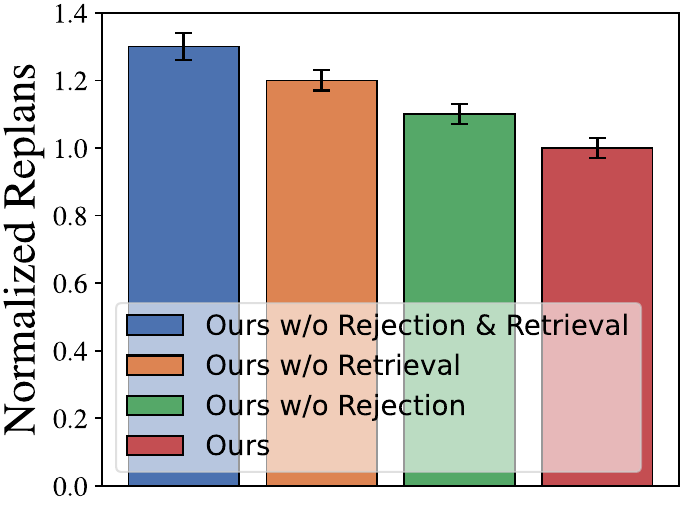}}}
        \vspace{-0.4cm}
        \caption[Rejection and retrieval modules]{\textbf{Rejection and retrieval.}
        Both the Rejection and Retrieval Modules improve performance, and combining them yields the best results.}
        \label{fig:ablation-1}
    \end{subfigure}
    \hfill
    \begin{subfigure}[t]{0.32\textwidth}
        \centering
        \makebox[\linewidth][c]{\raisebox{-0.917mm}[40mm][0pt]{\includegraphics[height=42mm]{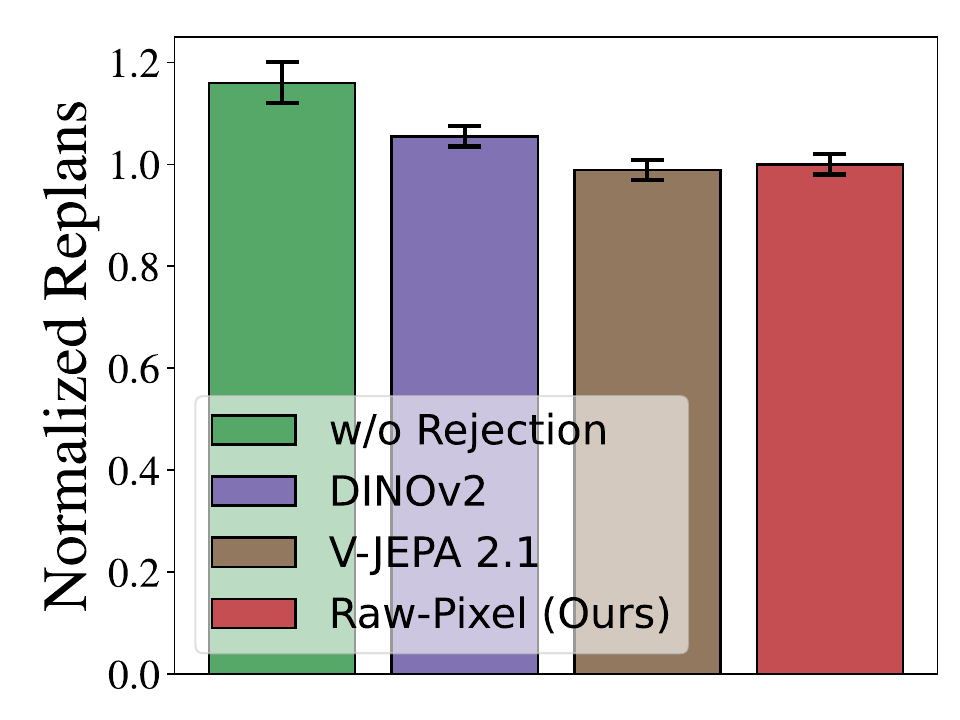}}}
        \vspace{-0.4cm}
        \caption[Rejection module distance]{\textbf{Rejection distance.}
        All three distances improve over no rejection; V-JEPA 2.1 is comparable to raw pixels.}
        \label{fig:ablation-3}
    \end{subfigure}
    \hfill
    \begin{subfigure}[t]{0.32\textwidth}
        \centering
        \makebox[\linewidth][c]{\raisebox{0mm}[40mm][0pt]{\includegraphics[width=\linewidth]{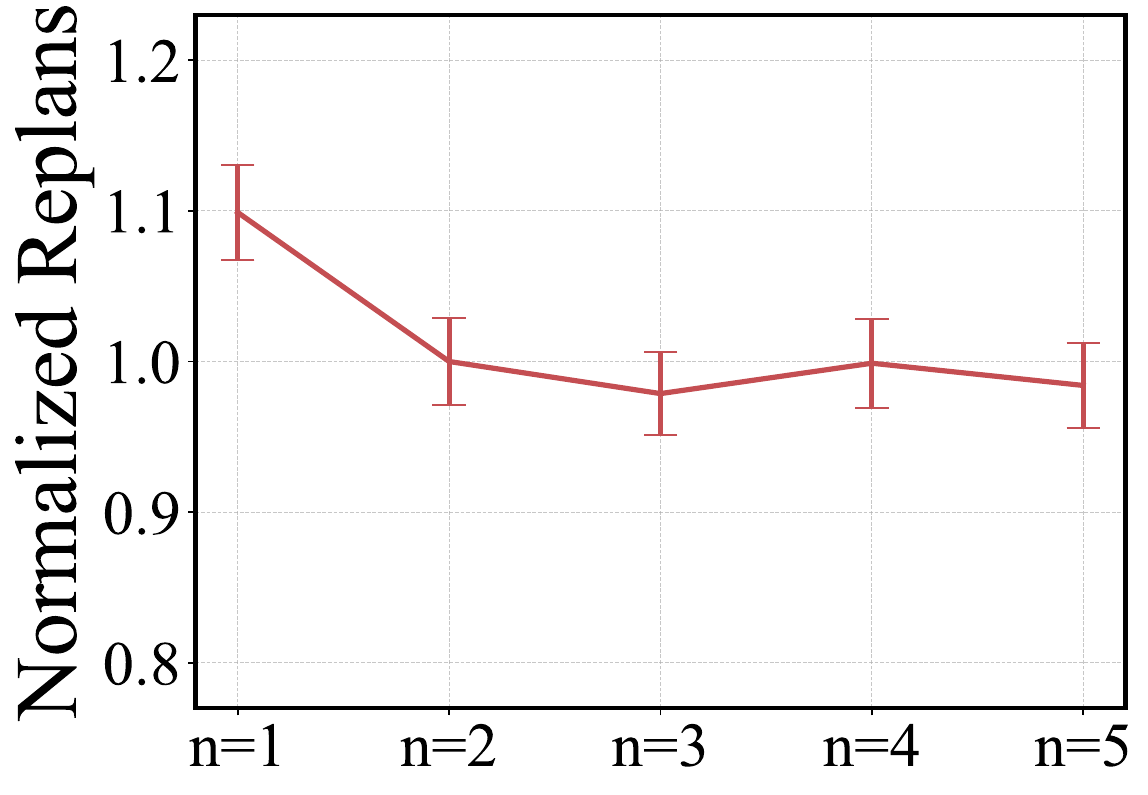}}}
        \vspace{-0.4cm}
        \caption[Number of generated video plans]{\textbf{Candidate plans.}
        We vary the number of video plans $n \in [1,5]$ and set $n=2$ to balance computational cost and performance gain.}
        \label{fig:ablation-2}
    \end{subfigure}

    \vspace{-1pt}
    \caption{\textbf{Ablation studies.}
    We evaluate variations of our method to assess the effectiveness of each component. The number of replans is normalized and aggregated across five tasks.}
    \label{fig:ablation}
\end{figure*}

\textbf{Evaluation metrics.}
\label{sec:exp:metric}
Our study focuses on two key aspects. First, we hypothesize that our replanning mechanism can reduce the number of replans required for successful execution. To measure the rate of adaptation, we report the average number of replans needed per successful attempt in our main experiment (\Cref{tab:mw_bench_envs}). Specifically, the agent must repeatedly replan until it either succeeds or reaches a predefined maximum number of replans, $M$. For all tasks, we set $M=14$.

Second, we assess the effectiveness of the replanning mechanism by directly comparing the replanned videos to ground-truth interaction videos. To quantify this similarity, we use standard video metrics: PSNR, LPIPS \cite{lpips}, \ICLRRevision{CLIP-score, and DINO-score.}

\textbf{Implementation and baselines.}
Across both suites, the Video Plan Generator is trained on videos in the offline experience dataset $\mathcal{D}$, whose size per suite is listed in \Cref{tab:dataset}. Each task is evaluated over 400 trials. At test time, $\mathcal{D}$ is stored as DINOv2 features rather than videos, at most 346\,MB per task. We compare our approach with AVDC, which replans without using past interactions, a VLM-based plan verifier, and imitation learning and offline reinforcement learning baselines.

\begin{itemize}
    \item \textbf{AVDC \citep{ko2024learning}.} The baseline with only the Action Module, \ie without the Retrieval and Rejection Modules introduced in this paper.
    \item \textbf{VLM verifier.} Inspired by VLMPC's VLM-guided evaluation of predicted futures~\cite{zhao2024vlmpc} and HAVE's history-aware candidate verification~\cite{li2025learn}, we construct a VLM verifier for our video-planning setting, which tests whether a general-purpose judge can replace $K_{rej}$. We generate candidate plans using the base video model without the canonical embedding and replace the Rejection Module with Gemini 3.1 Flash-Lite. The VLM receives the candidate video plans, the failed plan, and its corresponding failed interaction. It then scores the candidates, and we execute the best-scored plan.
\end{itemize}

We additionally compare against two state-embedding-conditioned policy learning baselines, which can be viewed as solving our problem with access to additional ground-truth action labels.
\begin{itemize}
    \item \textbf{BC \citep{pomerleau1989alvinn}.} A behavior cloning policy conditioned on the state and the retrieved embedding.
    \item \textbf{CQL \citep{kumar2020conservative}.} A conservative actor-critic where both the Q-function and policy are conditioned on the retrieved embedding.
\end{itemize}

\vspace{-5pt}
\subsection{Main Results}
\begin{table}[t]
    \caption{\textbf{Comparison with latent-context adaptation methods} on the Meta-World suite ($\downarrow$).
    Mean $\pm$ SEM replanning counts over 400 trials.
    The first pair of columns evaluates each method as a latent-context-conditioned actor-critic policy trained with dense rewards and action labels, which RELIC does not use.
    The second pair replaces our retrieval with the latent inferred by each method and keeps the rest of the pipeline unchanged.
    \textbf{Overall} is the mean per-task ratio to RELIC with its SEM, as in \Cref{tab:mw_bench_envs}.}
    \label{tab:latent_context}
    \centering
    \footnotesize
    \setlength{\tabcolsep}{1.5pt}
    \renewcommand{\arraystretch}{1.18}
    \begin{tabular*}{\columnwidth}{@{\extracolsep{\fill}}l*{5}{c}@{}}
    \toprule
    & \multicolumn{2}{c}{\scriptsize\emph{Latent-conditioned policy}} & \multicolumn{2}{c}{\scriptsize\emph{Latent in RELIC pipeline}} & \\
    \cmidrule(lr){2-3}\cmidrule(lr){4-5}
    \textbf{Task} & HiP-MDP & PEARL & HiP-MDP & PEARL & \textbf{RELIC} \\
    \midrule
    \pushbar & \replanentry{8.65}{0.24} & \replanentry{8.64}{0.23} & \replanentry{7.88}{0.24} & \replanentry{5.73}{0.21} & \replanentry{\mathbf{4.26}}{0.20} \\
    \pickbar & \replanentry{7.65}{0.23} & \replanentry{8.99}{0.24} & \replanentry{9.62}{0.23} & \replanentry{5.70}{0.21} & \replanentry{\mathbf{3.84}}{0.18} \\
    \slidebrick & \replanentry{11.29}{0.22} & \replanentry{13.20}{0.13} & \replanentry{10.35}{0.23} & \replanentry{7.69}{0.23} & \replanentry{\mathbf{6.69}}{0.26} \\
    \openbox & \replanentry{13.01}{0.15} & \replanentry{13.09}{0.14} & \replanentry{13.21}{0.13} & \replanentry{1.60}{0.04} & \replanentry{\mathbf{1.25}}{0.10} \\
    \turnfaucet & \replanentry{13.03}{0.14} & \replanentry{2.57}{0.09} & \replanentry{13.06}{0.14} & \replanentry{1.60}{0.05} & \replanentry{\mathbf{1.39}}{0.14} \\
    \midrule
    \textbf{Overall} & \replanentry{5.10}{0.26} & \replanentry{3.73}{0.18} & \replanentry{5.17}{0.26} & \replanentry{1.28}{0.04} & \textbf{1.00} \\
    \bottomrule
    \end{tabular*}
\end{table}

\vspace{-5pt}
\label{sec:exp:main-result}

The replanning efficiency of all methods is presented in \Cref{tab:mw_bench_envs}.
RELIC has lower mean replanning counts than AVDC, BC, CQL, and the VLM verifier on all eight tasks. Tasks with a discrete interaction mode, \openbox and \turnfaucet, are resolved within about one replan by most methods, so the performance gap is more apparent on tasks with continuous parameters. RELIC's remaining replans concentrate on friction and elasticity, \slidebrick and \textsc{Bounce}, where one failed interaction reveals less than a tilt or an overshoot does.

\textbf{Adapted plans are closer to ground truth.} In \Cref{tab:cv-metrics-left}, we report the evaluations of video plans generated by AVDC and our methods with various context encoders $E$, including R3M~\cite{nair2022rm}, CLIP~\cite{radford2021learning}, and DINOv2~\cite{oquab2024dinov}.  
Our method produces videos closer to the ground-truth videos than AVDC on all four metrics, while the replanning counts in \Cref{tab:mw_bench_envs} separate the methods far more sharply. 
Our method with DINOv2 outperforms its counterpart using R3M or CLIP. 

\textbf{Retrieval and refinement outperform latent-context inference.} Since our setting is closely related to hidden-parameter MDPs, we additionally compare against HiP-MDP~\cite{killian2017hipmdp} and PEARL~\cite{rakelly2019efficient}; because these methods differ from the video-planning baselines in supervision and architecture, we report them separately in \Cref{tab:latent_context}. We evaluate each method both as a latent-context-conditioned policy and as a drop-in replacement for our retrieval within the RELIC pipeline. RELIC requires fewer replans than all four variants on every task. The pipeline variants outperform the policy variants, and PEARL-RELIC is the strongest baseline; yet substituting PEARL's latent for our retrieval and refinement still costs more replans.

\textbf{Fewer replans outweigh the planning overhead.} Execution dominates the cost of each replanning round: a failed attempt in \pushbar{} takes about $25$\,s of wall-clock time before the next round can begin. Planning is a smaller overhead: on an H100 GPU, one round takes $3.5$\,s for RELIC versus $0.64$\,s for AVDC, with embedding refinement accounting for most of the difference. Because RELIC needs fewer replans, its expected time to first success is lower, about $121$\,s versus $175$\,s for AVDC.

\subsection{What Does the State Embedding Capture?}
\label{sec:exp:analysis}
We characterize the embedding on the Meta-World suite through three properties: it implicitly captures the hidden parameter $\theta$, it is the dominant determinant of success, and it generalizes beyond the configurations stored in $\mathcal{D}$.

\textbf{$\theta$ is decodable from the embeddings.} The probed embedding is a concatenation of frozen per-frame DINOv2 features; no component is trained on $\theta$. Although RELIC never observes $\theta$, the hidden parameters are recoverable from its state embeddings. For each Meta-World task, we fit a linear probe from the per-video embeddings $e_i$ to the ground-truth $\theta$, which is used for evaluation only. We use ridge regression for continuous parameters and logistic regression for discrete ones, with five-fold cross-validation grouped by configuration ID. As shown in \Cref{tab:theta_probe}, $\theta$ is linearly decodable from interaction embeddings on all five tasks. As a control, probing the same encoder's features of the \emph{pre-interaction} first frame yields $R^2$ below zero, worse than predicting the mean, and near-chance accuracy on 40 samples for the discrete tasks. The parameters become observable only through physical interaction, which suggests the embeddings capture interaction dynamics rather than scene appearance.

\textbf{An accurate $\theta$ hypothesis is sufficient for success.} Conditioning the planner on the ground-truth $\theta$, with the Action Module unchanged, yields first-attempt success on $88$--$100\%$ of trials across the five Meta-World tasks, with zero replans. Task success is thus governed by the planner's hypothesis about $\theta$, which is exactly what refinement updates.

\begin{figure}[t]
    \centering
    \includegraphics[width=\columnwidth]{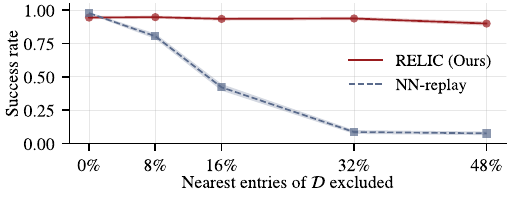}
    \caption{\textbf{Success rate on \pushbar{} as the entries of $\mathcal{D}$ nearest to the test $\theta$ are excluded.} NN-replay copies the action of the most similar episode retrieved from $\mathcal{D}$. 400 trials per point, 16-attempt cap.}
    \label{fig:bank_exclusion}
\end{figure}

\textbf{The embedding generalizes beyond stored configurations.} We stress-test RELIC by removing from $\mathcal{D}$ the configurations nearest to the test $\theta$ at inference time. As shown in \Cref{fig:bank_exclusion}, RELIC's success rate on \pushbar{} remains nearly flat as an increasing fraction of the entries of $\mathcal{D}$ nearest to the test $\theta$ is removed. A nearest-neighbor replay baseline that copies the action of the most similar episode in $\mathcal{D}$ is marginally higher than RELIC with the full dataset, when every test value has a close stored match, but collapses under the same exclusion while RELIC stays nearly flat: replay depends on dense coverage of $\theta$, whereas RELIC recovers $\theta$ from the interaction itself and remains accurate when no nearby configuration is stored.

\begin{table}[t]
    \caption{\textbf{Linear probes decoding the hidden parameter $\theta$ from state embeddings.}
    Five-fold cross-validation grouped by configuration ID, out-of-fold metrics.
    Continuous $\theta$ reports $R^2$; discrete $\theta$ reports accuracy.
    \emph{Post} probes embeddings of the interaction video; \emph{Pre} probes the same encoder's features of the first frame, before any physical contact.}
    \label{tab:theta_probe}
    \centering
    \footnotesize
    \setlength{\tabcolsep}{3pt}
    \renewcommand{\arraystretch}{1.12}
    \begin{tabular*}{\columnwidth}{@{\extracolsep{\fill}}lllcc@{}}
    \toprule
    \textbf{Task} & \textbf{Hidden $\theta$} & \textbf{Metric} & \textbf{Post} & \textbf{Pre} \\
    \midrule
    \pushbar & center of mass & $R^2$ & 0.77 & $-0.002$ \\
    \pickbar & center of mass & $R^2$ & 0.82 & $-0.002$ \\
    \slidebrick & friction & $R^2$ & 0.84 & $-5.5$ \\
    \openbox & push vs.\ pull & acc.\ (\%) & 100 & 65 \\
    \turnfaucet & CW vs.\ CCW & acc.\ (\%) & 92.5 & 55 \\
    \bottomrule
    \end{tabular*}
\end{table}

\subsection{Ablation Studies}
\label{sec:exp:ablation}

\paragraph{Retrieval initialization reduces replans}
We isolate the effect of retrieval-based initialization by comparing the full RELIC framework with RELIC (Refine FS), which refines state embeddings initialized by sampling from a multivariate Gaussian fit to the stored canonical embeddings, as described in \Cref{sec:method:planner}. As shown in \Cref{tab:mw_bench_envs}, the full framework requires fewer replans, while Refine FS yields competitive performance and \emph{does not require access to $\mathcal{D}$ at test time}. 

\paragraph{Removing either module costs replans}

To investigate the effectiveness of the Rejection and Retrieval Modules, we compare our full method against variants that remove either the Rejection or the Retrieval component. The result in \Cref{fig:ablation-1} shows the normalized number of replans aggregated across the five Meta-World tasks. Removing either component consistently increases the number of replans. 

\paragraph{Distance choice matters little}
We compare raw-pixel, DINOv2, and V-JEPA 2.1 distances for rejection in \Cref{fig:ablation-3}. All three reduce normalized replans relative to disabling rejection. Despite using video representations, V-JEPA 2.1 performs comparably to raw pixels, with overlapping error bars. These results support flexibility in the distance function, and we use raw-pixel L2 distance, which requires no encoder. \Cref{sec:apdx:rejection_metric} tests robustness to background noise and segmentation.

\begin{figure*}[t]
    \centering
    \includegraphics[clip, trim=0cm 3.4cm 64.3cm 0cm, width=0.97\linewidth]{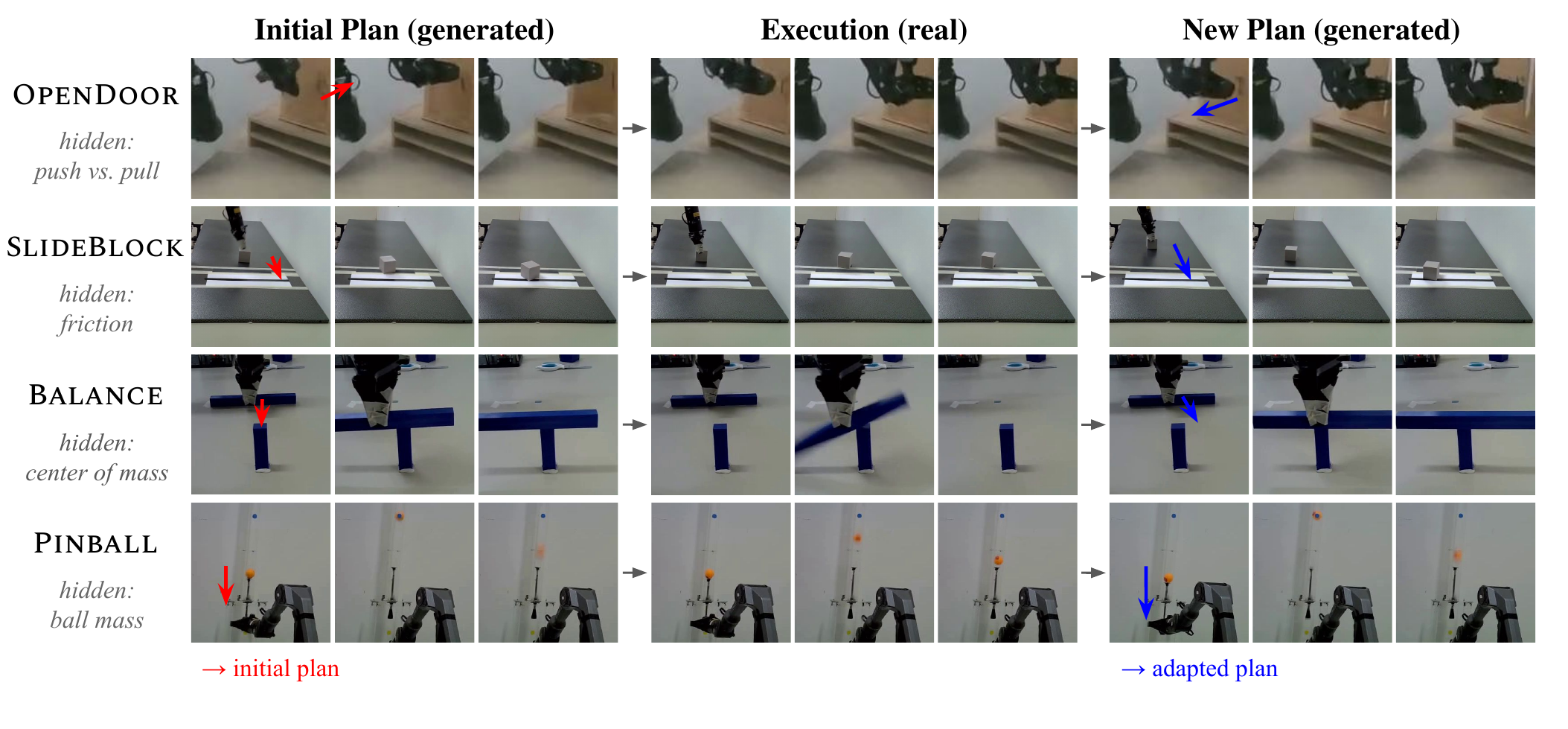}
    \vspace{-0.4cm}
    \caption{\textbf{Real-world qualitative results.} Each row is one adaptation trial on a real-world task whose hidden parameter is noted under the task name. The initial plan reflects a wrong hypothesis about that parameter, so its execution fails; RELIC then replans from the failed interaction, and the adapted plan succeeds when executed, as marked in the column headers. Arrows indicate the planned action or placement in the {\color{red}initial} and {\color{blue}adapted} plans; they are for visualization only and not part of the system input.}
    \label{fig:real_qual}
\end{figure*}

\paragraph{Two candidates capture most of the gain}
As shown in \Cref{fig:ablation-2}, increasing the number of candidate plans from $n=1$ to $n=2$ reduces replans by approximately $10\%$ across the five Meta-World tasks. By having two candidates, the Rejection Module can favor an alternative to previous failures. Further increases to $n=3,4,5$ yield diminishing returns. We therefore retain $n=2$ to balance computational cost and replanning performance.

\paragraph{Why does raw-pixel rejection suffice?}
Together, \Cref{fig:ablation-3,fig:ablation-2} suggest that the operative mechanism is elimination: distances between generated plans reliably flag candidates that resemble past failures, whereas ranking the remaining candidates by hidden physical differences remains open. VLMs could offer more flexible interpretation of failed plans and observed interactions. However, in the tested configuration, our VLM verifier still requires more replans on average than RELIC in \Cref{tab:mw_bench_envs}. Developing feedback that more reliably connects visual discrepancies to hidden physical variation, through learned distances or VLM-based reasoning, is a promising direction for future work.

\subsection{Real-world Evaluation}
\label{sec:exp:real}

\textbf{Tasks.} We evaluate on four real-world robotic manipulation tasks that require inferring hidden physical properties not observable from appearance alone. These tasks span both discrete latent variables, such as push versus pull, and continuous ones, such as friction, mass, and center of mass, and require physical reasoning under partial observability.

\begin{itemize}
    \item \textbf{\opendoor}: 
    A door can be opened by \textit{pushing or pulling}, with the correct action determined by a hidden lock mechanism behind the door that is not observable to the robot. Success: the door opens by at least $10^{\circ}$.
    \item \textbf{\slideblock}: 
    The robot slides a block along a track to a target location, where the block’s \textit{friction coefficient} varies across trials, which directly affects the resulting displacement. Success: the block comes to rest overlapping the target line.
    \item \textbf{\balance}: 
    The robot places a bar on a pivot to maintain a level orientation, which requires estimating the bar’s unknown \textit{center of mass}. Success: the bar stays on the pivot after release.
    \item \textbf{\pinball}: 
    The robot pulls back a spring-loaded launcher to strike a ball toward a target height, requiring inference of the ball’s \textit{mass and internal weight distribution}, which nonlinearly affect the dynamics; for instance, the lightest ball requires the strongest impulse due to dominant air resistance. Success: the ball reaches a marker $22$\,cm above its rest position.
\end{itemize}

These tasks are performed across two robot platforms, an ALOHA WidowX arm for \opendoor and a Piper-X arm for the remaining three. We visualize the tasks in \Cref{fig:real_qual}; hardware, success criteria, and trial sampling are in \Cref{sec:apdx:real:task_hardware}.

\begin{table}[!htbp]
  \centering
  
  \caption{\textbf{Real-world success rates.} Successes out of trials for each task; \textit{Overall} reports the macro-averaged success rate. RELIC adapts to hidden interaction modes from a small number of trials.}
  \label{tab:realworld}
  \footnotesize
  \setlength{\tabcolsep}{1.5pt}
  \renewcommand{\arraystretch}{1.16}
  \begin{tabular*}{\linewidth}{@{\extracolsep{\fill}}lccccc@{}}
    \toprule
    \textbf{Method} & \opendoor & \slideblock & \balance & \pinball & \textbf{Overall} \\
    \midrule
    AVDC & 8/20 & 4/20 & 7/20 & 5/20 & 30.0\% \\
    \textbf{RELIC (Ours)} & \textbf{14/20} & \textbf{7/20} & \textbf{17/20} & \textbf{13/20} & \textbf{63.8\%} \\
    \bottomrule
  \end{tabular*}
\end{table}

\textbf{Evaluation.} For each task, we collect 18--25 real-world training videos. To maintain plan diversity across replanning rounds at this data scale, we inject pixel-level Gaussian noise with $\sigma=0.1$ into the first-frame observation at inference time, applied identically to RELIC and AVDC; \Cref{sec:apdx:real:hyperparam} reports the noise ablation.
 We evaluate the single-episode replanning success rate conditioned on an initial failure, comparing against AVDC, which replans without using past interactions. As shown in \Cref{tab:realworld}, RELIC improves over AVDC on every task. The largest gains come from \balance{} and \pinball{}, where a single failed attempt provides a strong physical signal: a tilt direction, or an over- or under-shot. \slideblock{} remains the hardest task for both methods due to its larger state distribution: the block can slide sideways to arbitrary positions, making accurate trajectory modeling more difficult.

\vspace{-0.1cm}
\FloatBarrier
\section{Limitations}
\label{sec:limitations}
To isolate the challenge of adapting the planner from interaction videos, our work assumes a reliable Action Module and does not adapt it; even with ground-truth $\theta$, up to $12\%$ of trials still fail on some tasks, an execution floor that adapting the planner alone cannot remove. Future work could jointly adapt the low-level policy alongside the planner. RELIC also does not take actions to expose hidden properties: every attempt aims to complete the task, and adaptation uses only the information that attempt happens to reveal. Active system-identification methods instead choose informative probing actions before committing~\cite{ljung1998system, memmel2024asid}, at the cost of specifying which parameters to estimate. Finally, properties that leave no visual trace in the interaction lie outside what the state embedding can capture and would require multi-modal sensing, such as tactile or proprioceptive feedback.

\section{Conclusion}
\label{sec:discussion}
We presented RELIC, a video planning framework that adapts to hidden physical properties from test-time failures by refining a latent embedding from past interactions and selecting plans that avoid past failures. Linear probes confirm that this embedding captures the hidden parameters from the interaction itself rather than from scene appearance. Across the simulated and four real-world tasks, RELIC improves over the baselines. Our real-world tasks further target diverse settings for hidden-parameter inference, including spring-loaded impact dynamics, surface friction, and center-of-mass estimation. We hope these results and the proposed evaluation encourage further work on video planning and adaptation.

\section*{Acknowledgment}
This work was supported by the Industrial Technology Research Institute, Taiwan, the National Science and Technology Council, Taiwan (114-2628-E-002 -021-), and the Taiwan Centers of Excellence.
Shao-Hua Sun was supported by the Yushan Fellow Program by the Ministry of Education, Taiwan.
\bibliographystyle{IEEEtranN}
\bibliography{references}  %

\clearpage
\useRomanappendicesfalse
\appendices
\crefalias{section}{appendix}
\crefalias{subsection}{appendix}

\section{Supplementary materials}
\label{sec:apdx:supp}

\subsection{Website and Code}
\label{sec:apdx:website}
Qualitative results and videos are on our \href{\ourwebsite}{website}; code is released at \url{https://github.com/RELIC-video/relic}.

\section{Benchmark Details of the Meta-World Suite}
\label{sec:apdx:benchmark}

\begin{figure*}[ht]
    \centering
    \begin{minipage}[b]{0.19\linewidth}
        \centering
        \includegraphics[width=\linewidth]{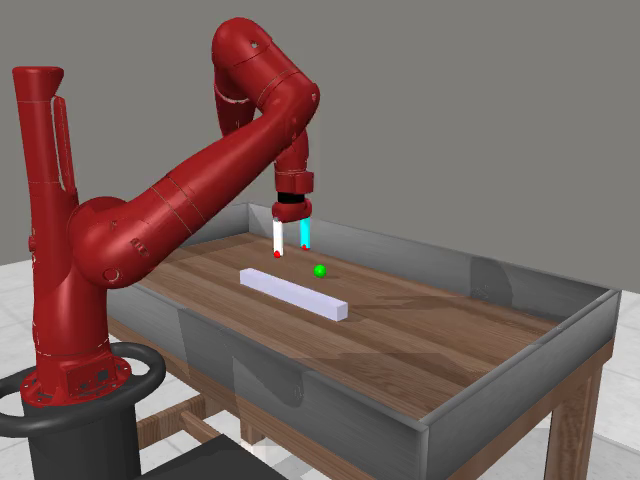}
        \pushbar
    \end{minipage}
    \begin{minipage}[b]{0.19\linewidth}
        \centering
        \includegraphics[width=\linewidth]{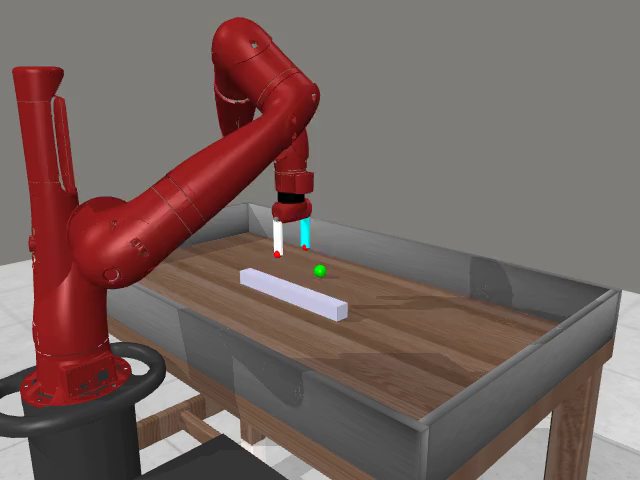}
        \pickbar
    \end{minipage}
    \begin{minipage}[b]{0.19\linewidth}
        \centering
        \includegraphics[width=\linewidth]{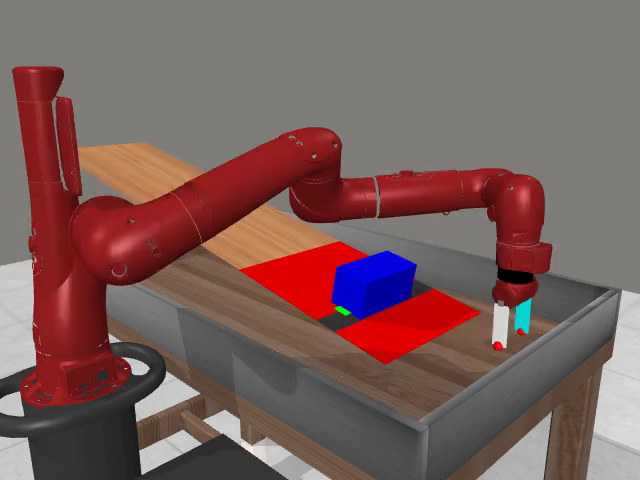}
        \slidebrick
    \end{minipage}
    \begin{minipage}[b]{0.19\linewidth}
        \centering
        \includegraphics[width=\linewidth]{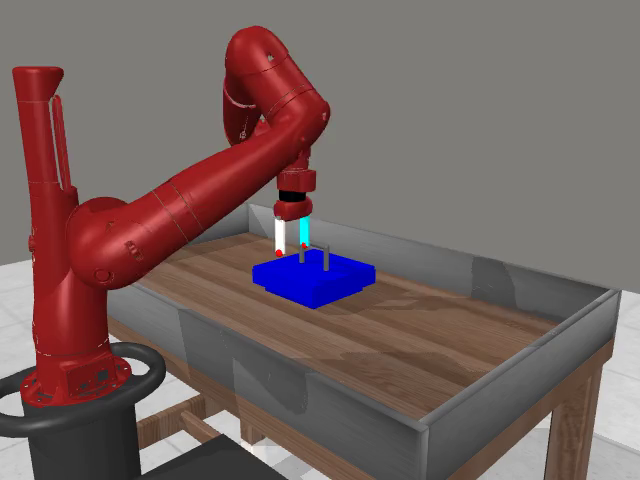}
        \openbox
    \end{minipage}
    \begin{minipage}[b]{0.19\linewidth}
        \centering
        \includegraphics[width=\linewidth]{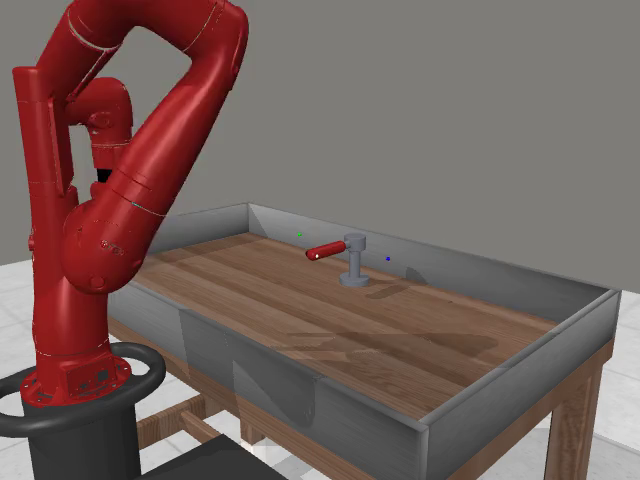}
        \turnfaucet
    \end{minipage}
    \caption{Settings of five task environments.}
    \label{fig:five-tasks}
\end{figure*}

The environments of the five tasks are shown in \Cref{fig:five-tasks}.

\subsection{\pushbar}

The objective of this task is to push a bar towards the target. Physically, the bar is combined with a mass point and a mass-free bar. Therefore, if the robot does not push the bar exactly at the center of mass, torque is imposed on the bar, resulting in the rotation of the bar. 

The bar has a length of 0.4 meters. For ease of data collection, we discretize the center of mass position during pretraining. We evaluate at 25 evenly spaced values in $[-0.18, 0.18]$ with step $0.015$, i.e., $\{-0.18, -0.165, \dots, 0.165, 0.18\}$. During evaluation, the C.M is randomly sampled from $[-0.18, 0.18]$.

The visualization of this task is shown in \Cref{figs:push-bar-video}.

\subsection{\pickbar}

The objective of this task is to grasp a bar towards the target. Physically, like the push-bar mentioned above, the bar is combined with a mass point and a mass-free bar. Therefore, if the robot does not grasp the bar exactly at the center of the mass, torque is imposed on the bar, resulting in the tilt of the bar.

The properties of the bar in this task are the same as the bar in push-bar. And the distances between the center of mass and the geometric center of the bar in the pretraining stage are also the same as above. 

The visualization of this task is shown in \Cref{figs:pick-bar-video}.

\subsection{\slidebrick}

There are two stages in this task. In the first stage, the robot pushes a brick up an inclined surface. In the second stage, the brick slides freely down the slope. The objective is to push the brick at a proper height so that the brick can slide down and stop at the gray band. 

Note that the wooden slope has a relatively low friction coefficient and that the red slope has a relatively high friction coefficient. Therefore, the brick can slide down and accelerate on the wooden slope and decelerate on the red slope. The friction coefficient of the wooden slope is fixed and is set to $0$, while the friction coefficient of the red slope varies. \ICLRRevision{During training, we discretize the friction coefficient as $\mu \in \{0.24, 0.25, 0.26, \dots, 0.40\}$ in our dataset $\mathcal{D}$. During evaluation, the friction coefficient is sampled from $[0.24, 0.40]$.} 

The visualization of this task is shown in \Cref{figs:slide-brick-video}.

\subsection{\openbox}

The objective of this task is to open a box whose lid operates in one of two possible ways: it either needs to be lifted or slid open. The specific mode is chosen at random and cannot be identified visually, so the robot must physically interact with the box and adjust its strategy based on the observed response. 

The visualization of the task is shown in \Cref{figs:open-box-video}.

\subsection{\turnfaucet}

The objective of this task is to rotate a faucet. The direction of rotation—clockwise or counterclockwise—is randomly assigned and cannot be determined in advance through observation. The robot must interact with the faucet to identify the correct direction and execute the appropriate motion. 

The visualization of the task is shown in \Cref{figs:turn-faucet-video}.

\begin{center}
    \begin{minipage}{\linewidth}
        \centering
        \includegraphics[width=0.95\linewidth]{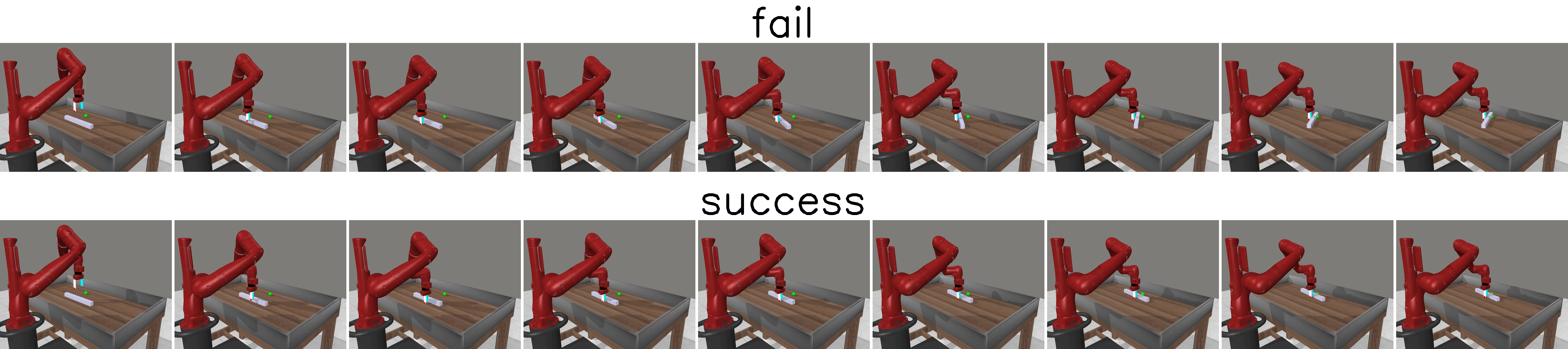}
        \captionof{figure}{\pushbar. The above video shows a failed trial because the bar's deflection is too large, preventing it from reaching the target.}
        \label{figs:push-bar-video}
    \end{minipage}

    \vspace{6pt}
    \begin{minipage}{\linewidth}
        \centering
        \includegraphics[width=0.95\linewidth]{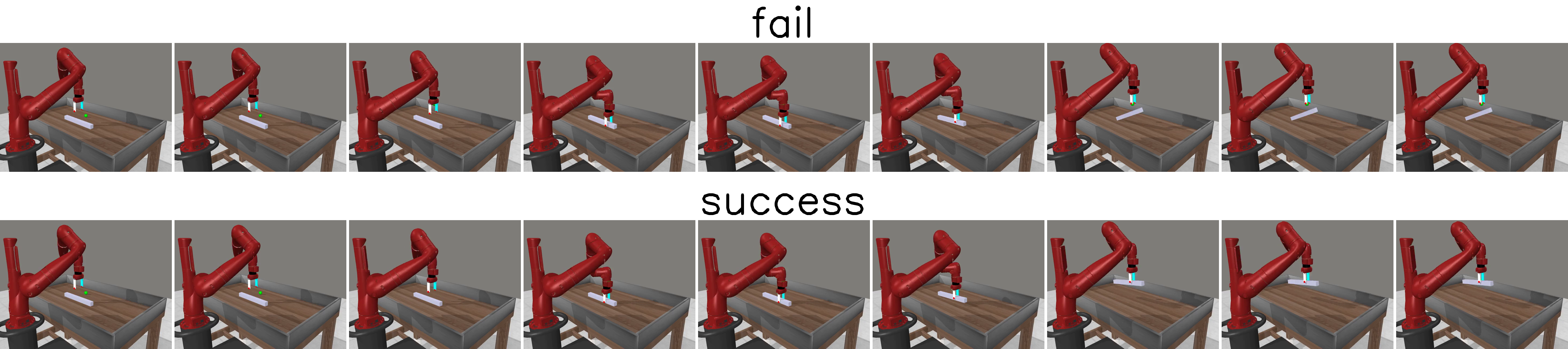}
        \captionof{figure}{\pickbar. The above video shows a failed trial because the bar fell.}
        \label{figs:pick-bar-video}
    \end{minipage}

    \vspace{6pt}
    \begin{minipage}{\linewidth}
        \centering
        \includegraphics[width=0.95\linewidth]{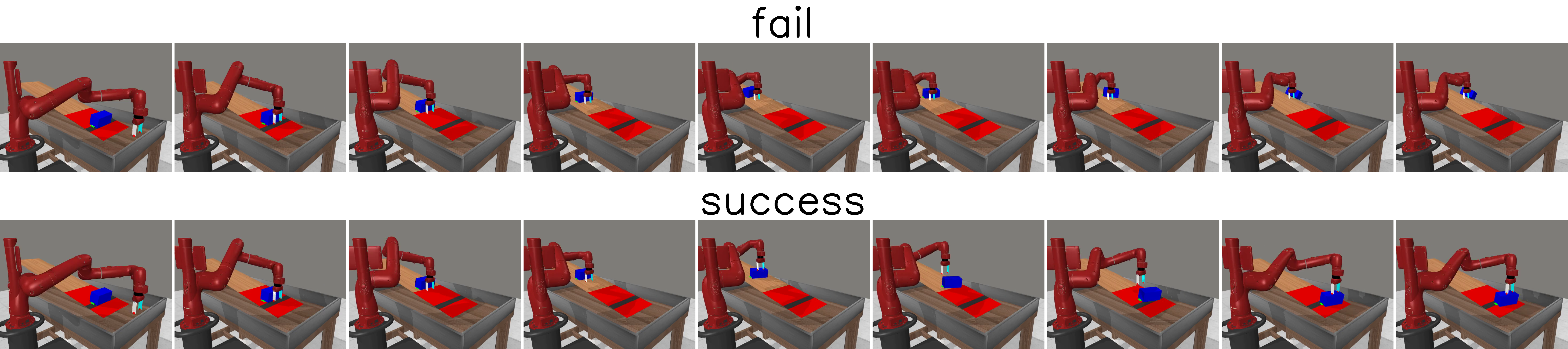}
        \captionof{figure}{\slidebrick. The above video shows a failed trial because the brick didn't slide down to the gray band.}
        \label{figs:slide-brick-video}
    \end{minipage}

    \vspace{6pt}
    \begin{minipage}{\linewidth}
        \centering
        \includegraphics[width=0.95\linewidth]{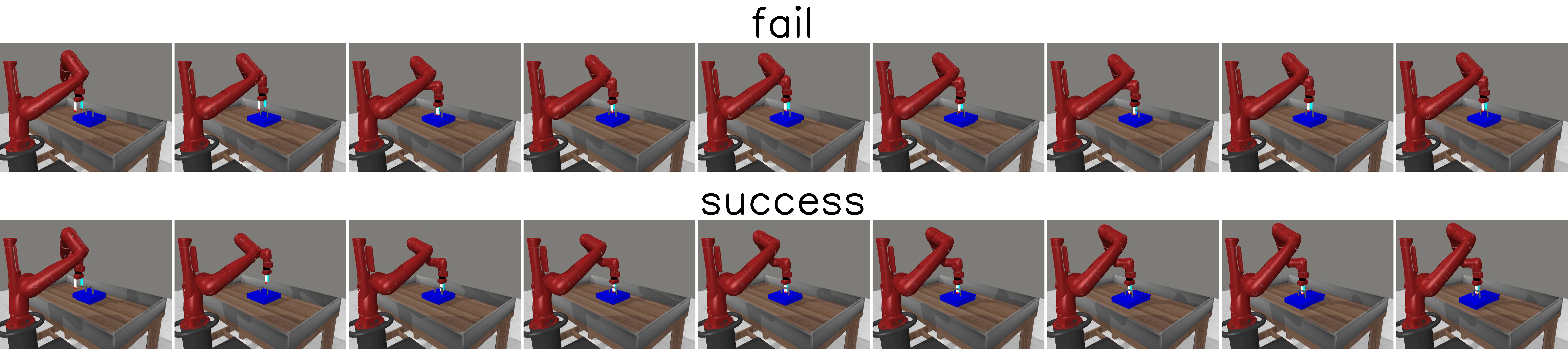}
        \captionof{figure}{\openbox. The above video shows a failed trial because the robot tries to grasp the handle upward; the correct way is to push the handle horizontally.}
        \label{figs:open-box-video}
    \end{minipage}

    \vspace{6pt}
    \begin{minipage}{\linewidth}
        \centering
        \includegraphics[width=0.95\linewidth]{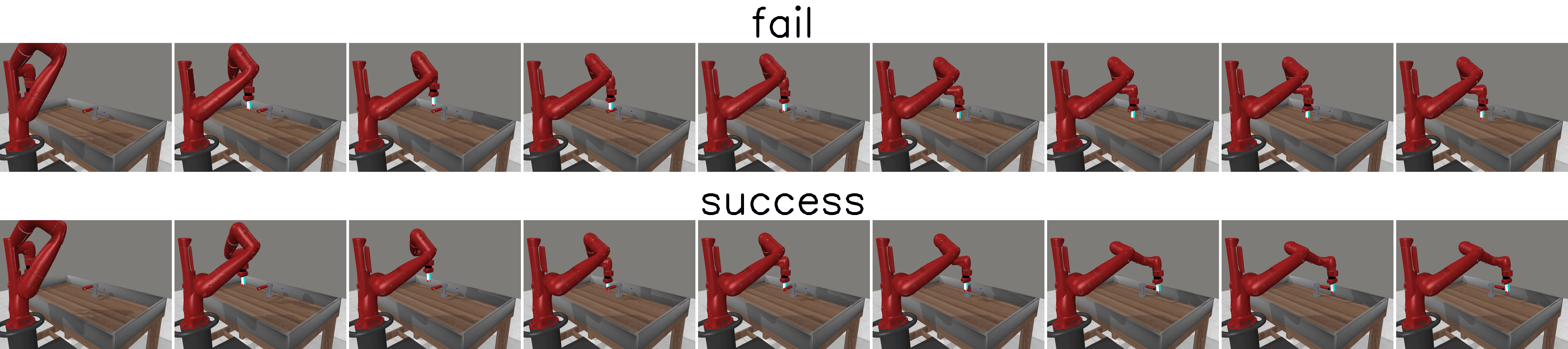}
        \captionof{figure}{\turnfaucet. The above video shows a failed trial because the robot tries to turn the faucet clockwise; the correct way is counterclockwise.}
        \label{figs:turn-faucet-video}
    \end{minipage}
\end{center}

\section{Real-world experiment}
\label{sec:apdx:real}

\subsection{Tasks and Hardware Setup}
\label{sec:apdx:real:task_hardware}

\begin{figure*}[t]
    \centering
    \setlength{\tabcolsep}{2pt}
    \renewcommand{\arraystretch}{1.0}

    \begin{tabular}{@{}cccc@{}}
        \multicolumn{2}{c}{\textbf{\opendoor}} &
        \multicolumn{2}{c}{\textbf{\slideblock}} \\[2pt]

        \includegraphics[width=0.24\linewidth]{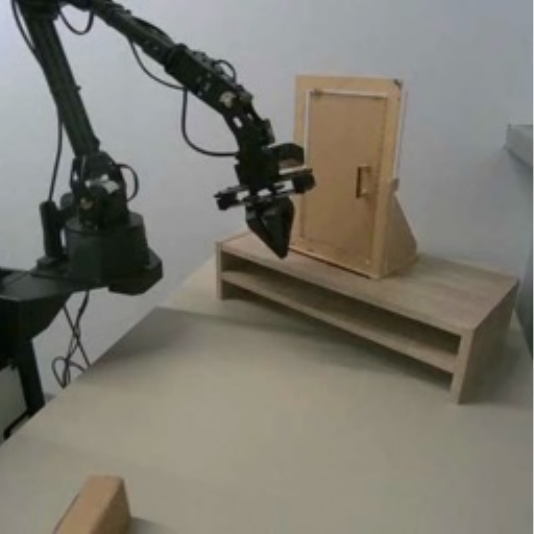} &
        \includegraphics[width=0.24\linewidth]{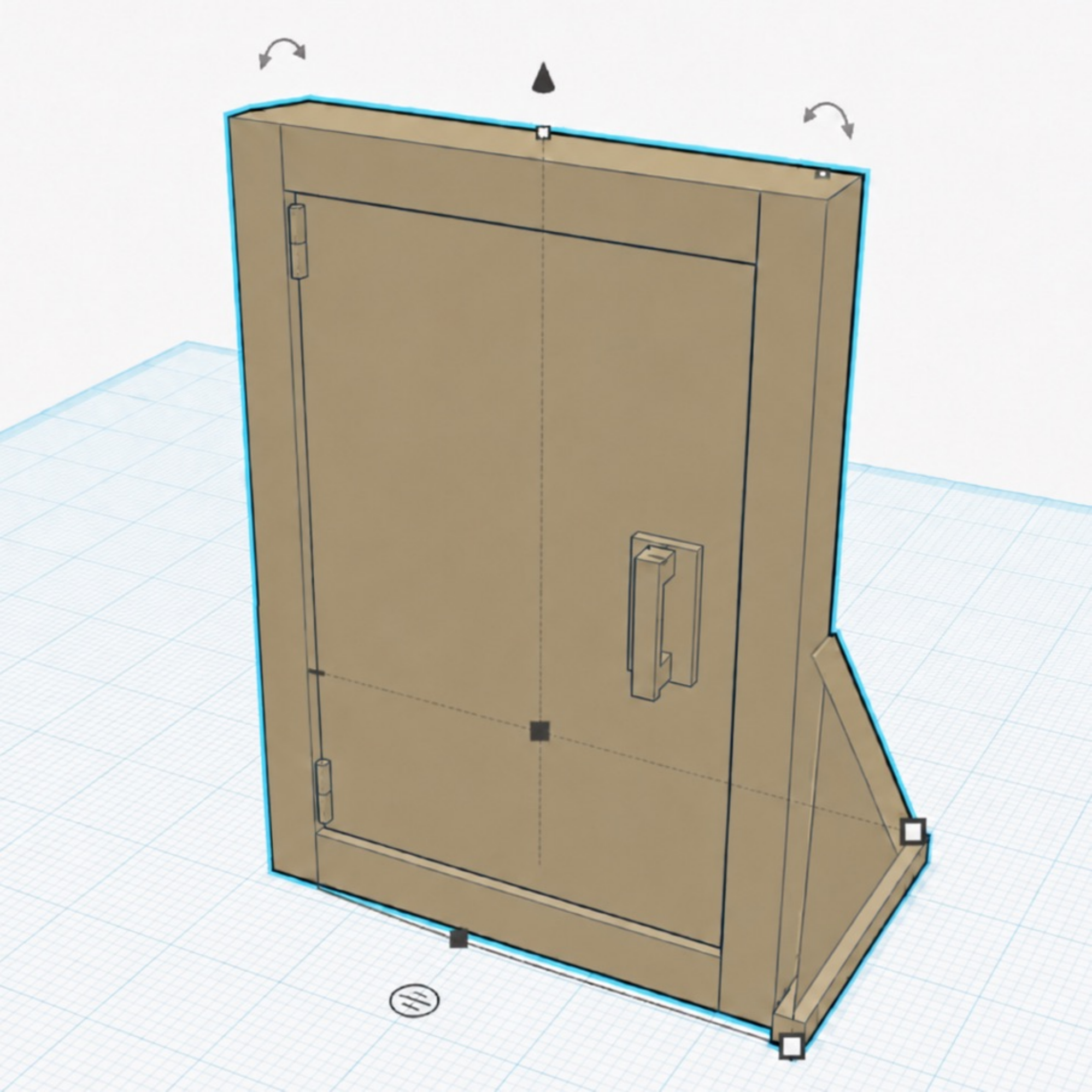} &
        \includegraphics[width=0.24\linewidth]{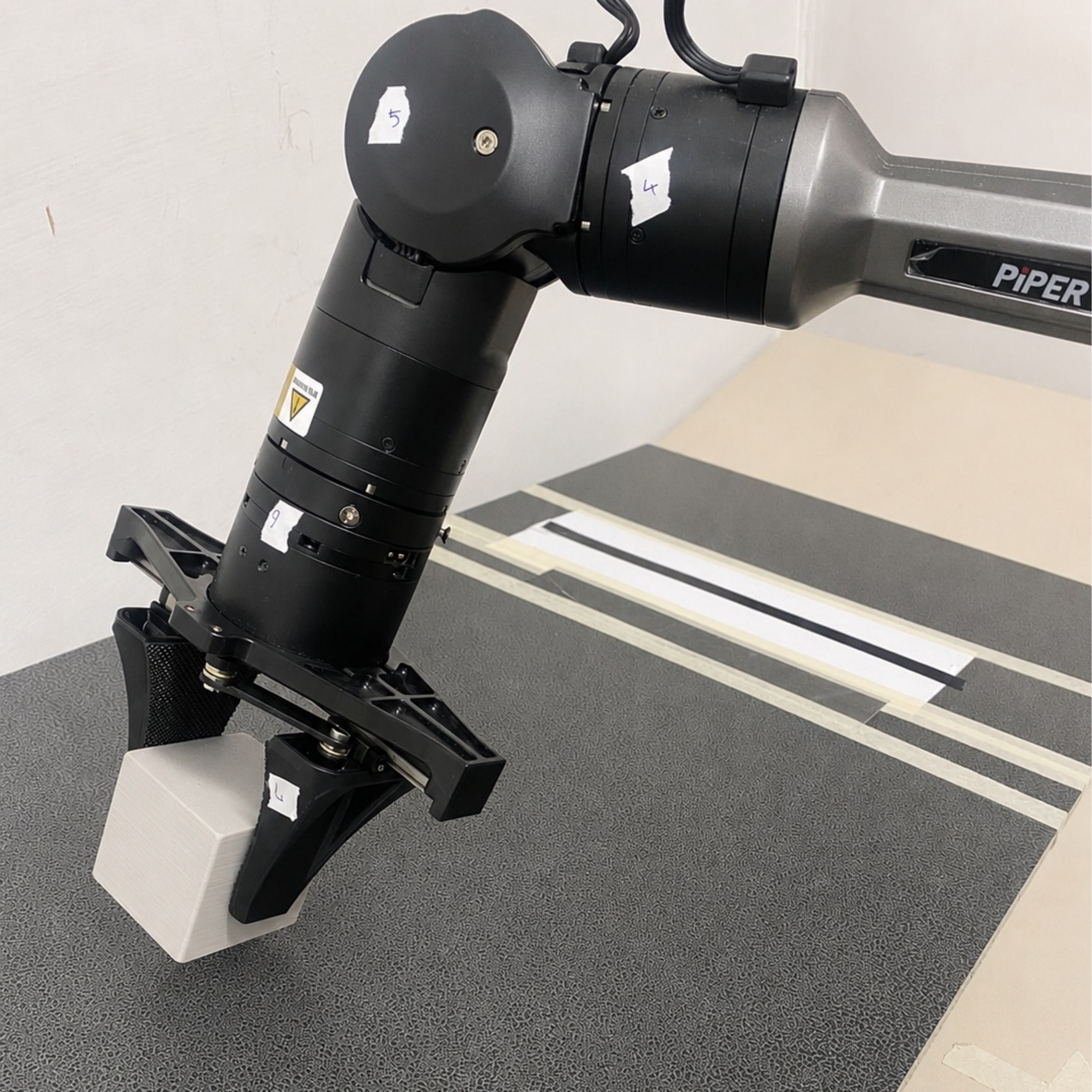} &
        \includegraphics[width=0.24\linewidth]{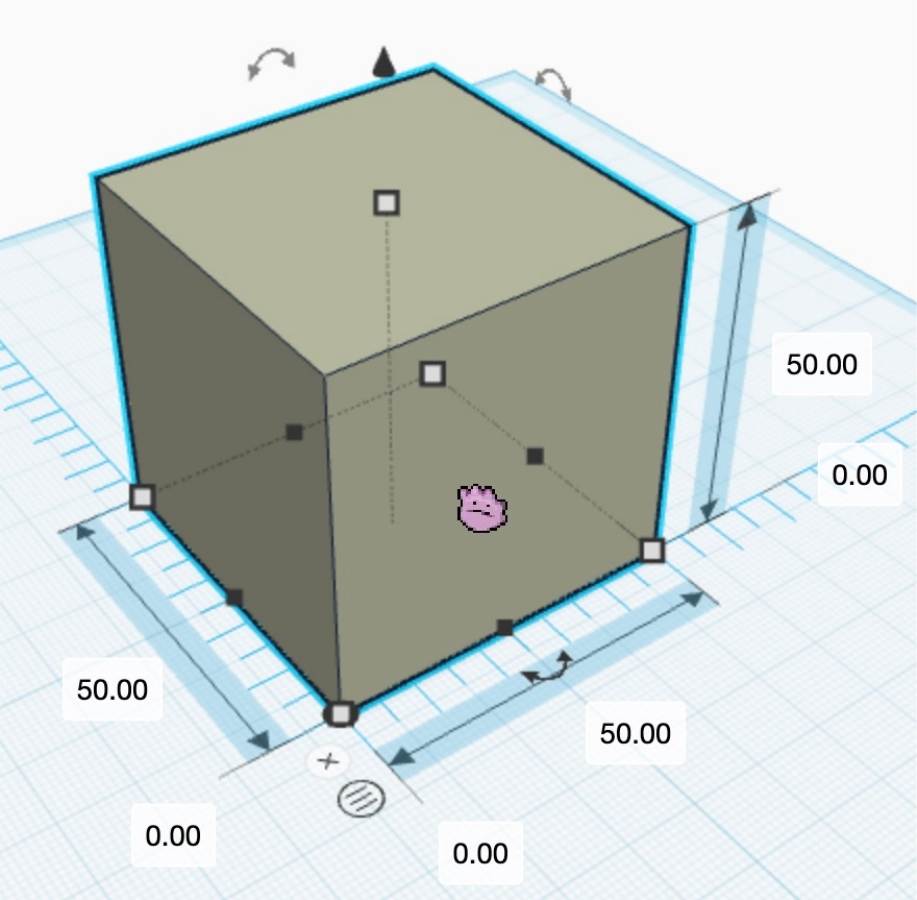} \\[-1pt]

        \multicolumn{2}{c}{\textbf{\balance}} &
        \multicolumn{2}{c}{\textbf{\pinball}} \\[2pt]

        \includegraphics[width=0.24\linewidth]{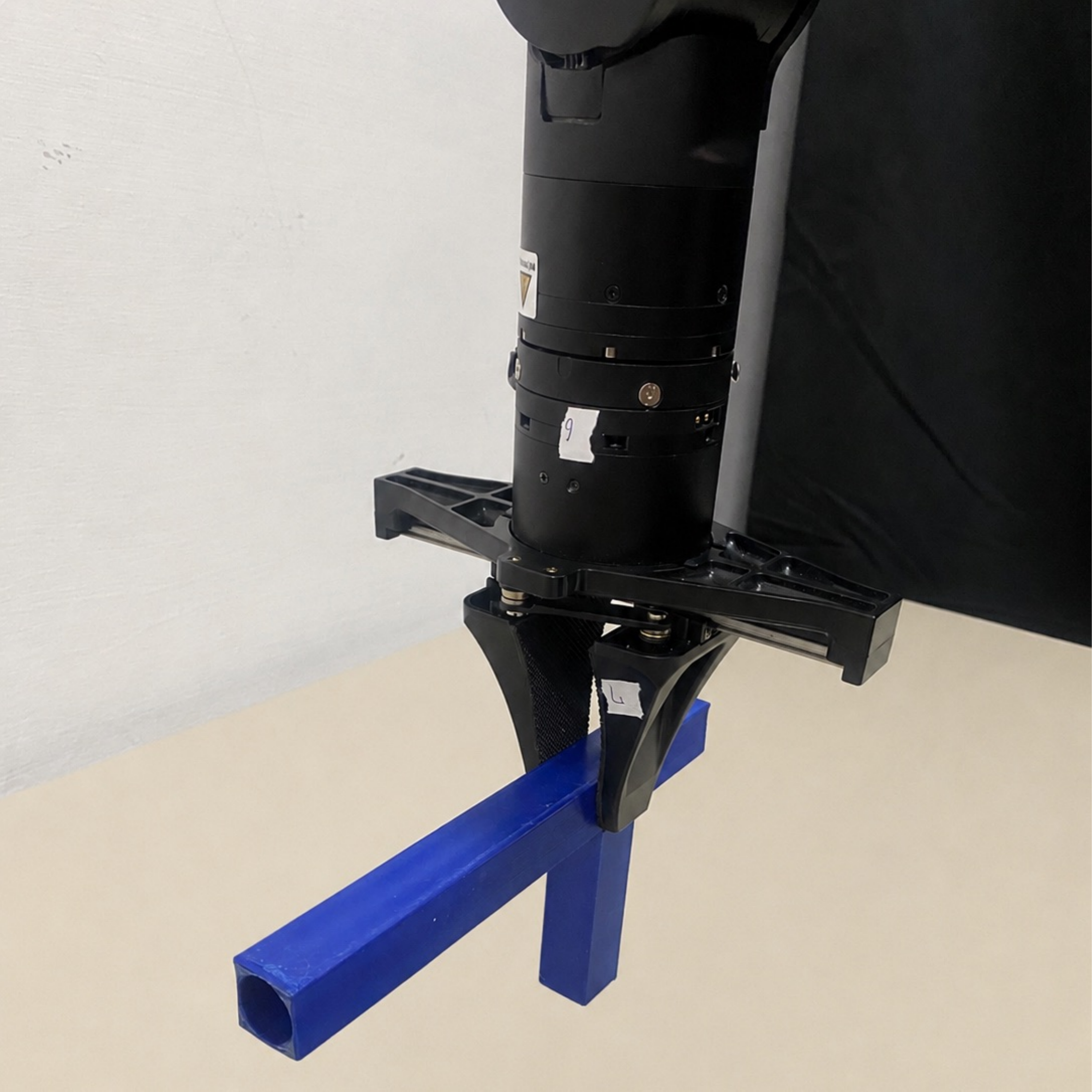} &
        \includegraphics[width=0.24\linewidth]{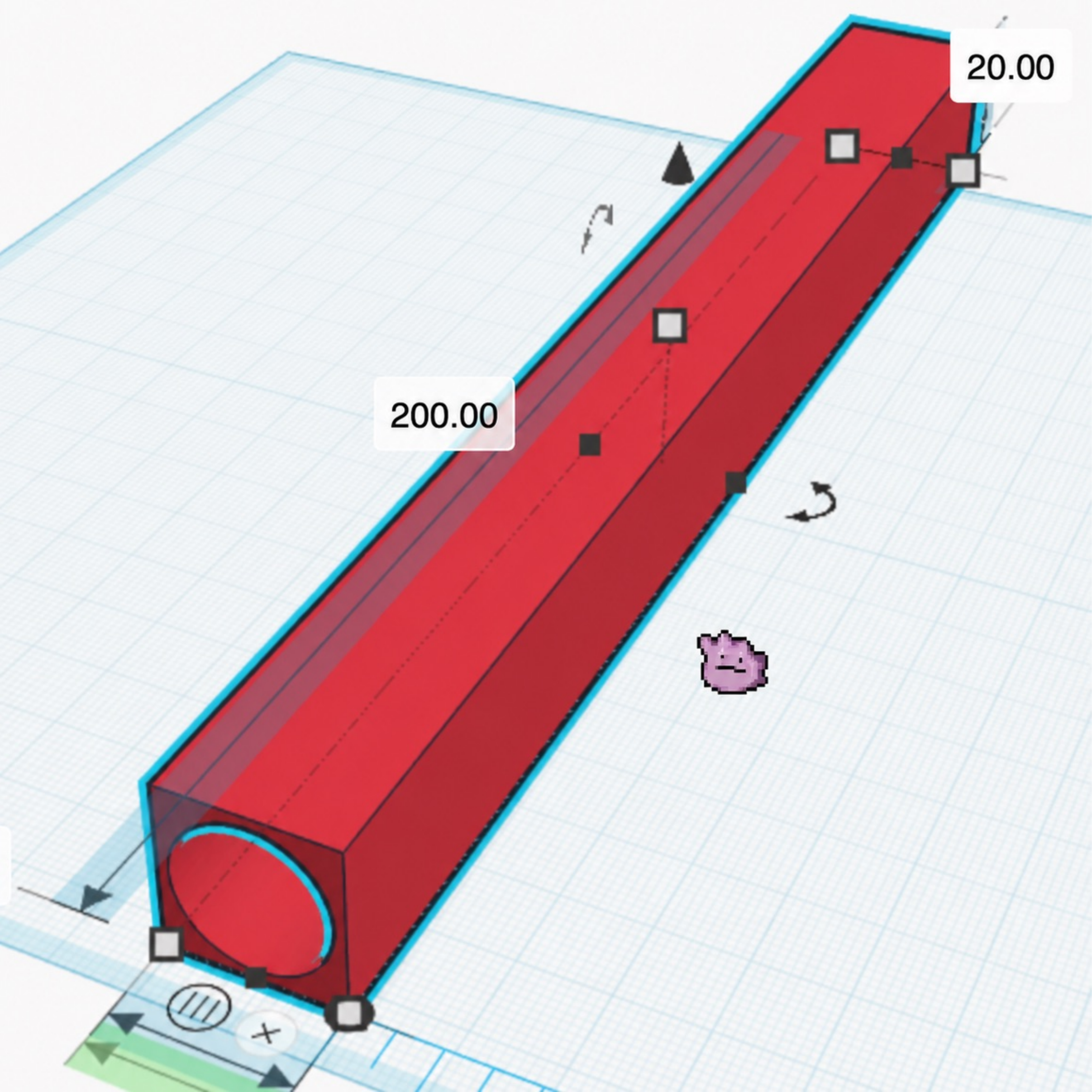} &
        \includegraphics[width=0.24\linewidth]{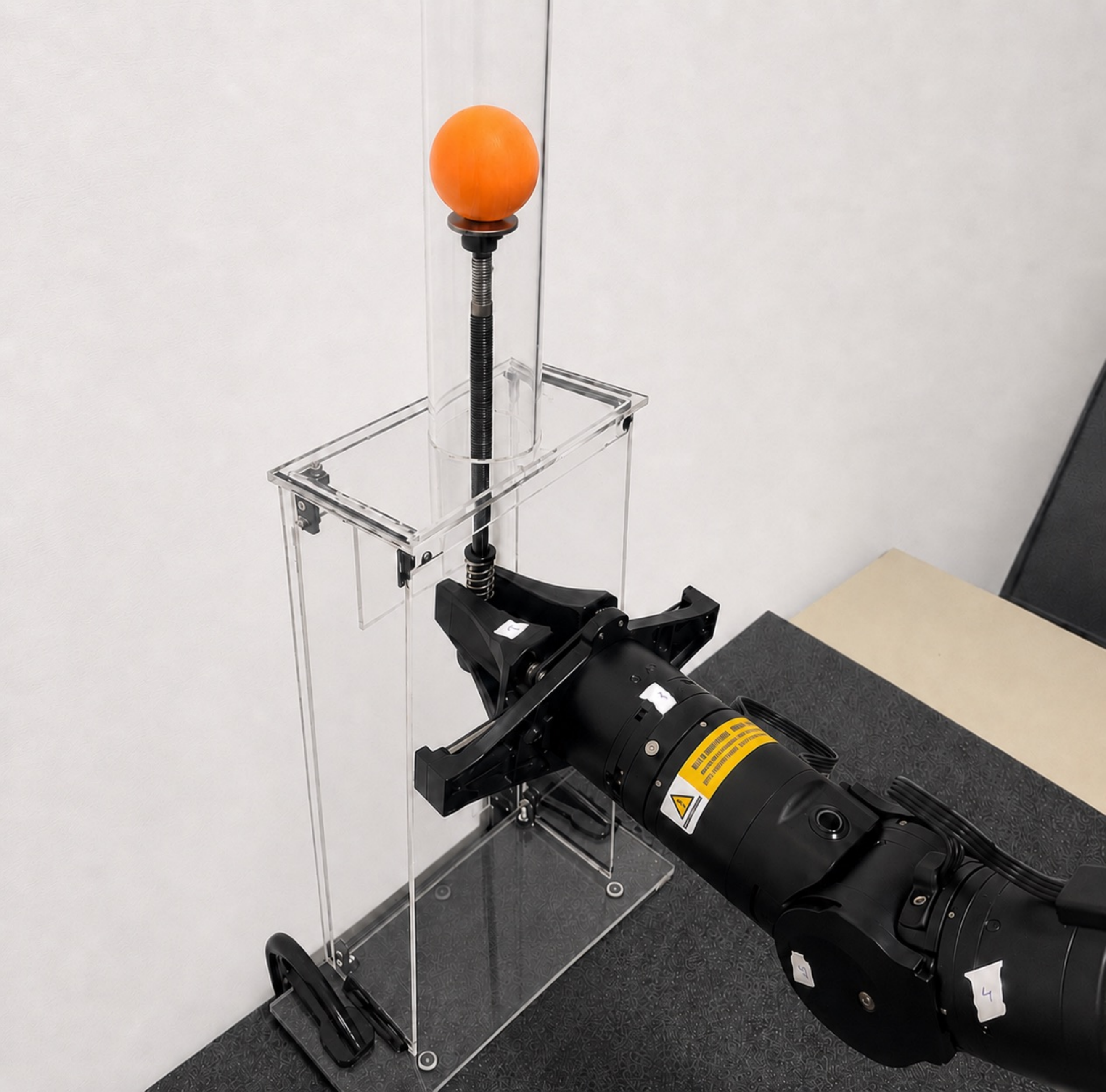} &
        \includegraphics[width=0.24\linewidth]{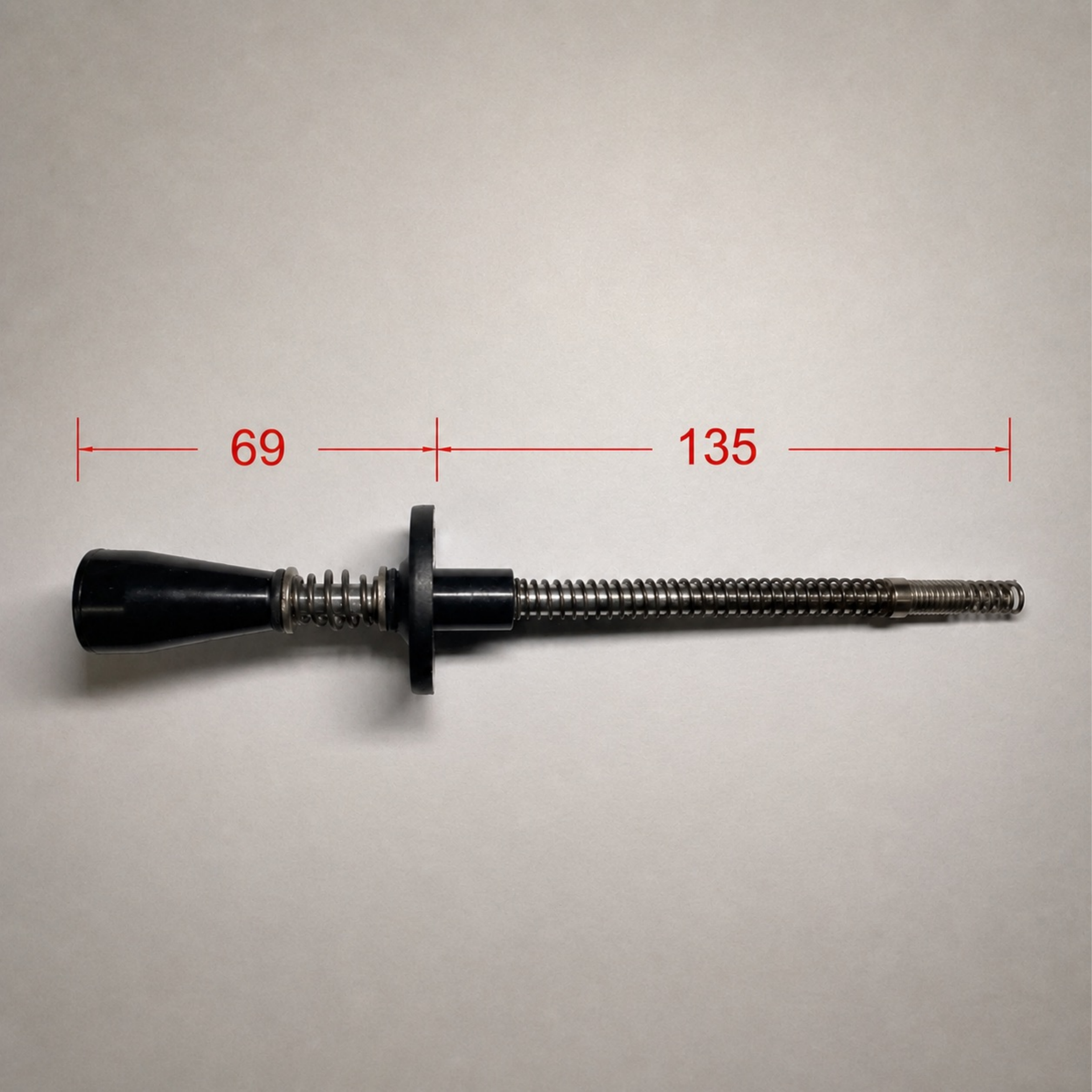} \\[-1pt]

    \end{tabular}

    \caption{\textbf{Real-world task setups.}
    For each task, we show the overall experimental setup and a detailed
view of the task-specific physical design. Dimensions shown in the
detail views are reported in millimeters. The Open Door detail is a
schematic illustration of the task mechanism and is not drawn to scale.
The hidden parameters correspond to push-versus-pull direction for
\opendoor, block friction for \slideblock, block center of mass for \balance,
and ball mass for \pinball.}
    \label{fig:realworld_tasks}
\end{figure*}

We evaluate on four real-world manipulation tasks, each containing a
hidden physical parameter that cannot be identified from appearance alone.
\Cref{fig:realworld_tasks} presents the overall experimental setup and
task-specific details for each task, while \Cref{tab:realworld_tasks}
summarizes the hidden parameter, robot platform, and data collection
statistics. All visual observations are captured with a single Logitech
C270 webcam.

\paragraph{\opendoor.}
The robot must open a cabinet door, which requires approaching the handle from a fixed home pose and actuating the door in the correct direction. A trial is successful if the door opens by at least 10 degrees; in practice, the door either swings open cleanly when actuated in the correct direction or jams immediately when actuated in the wrong direction, so partial successes do not occur. A smart policy should commit to one direction on the first attempt and, upon observing that the door jams, switch to the opposite direction on the next attempt; the binary nature of the parameter means a single failed interaction is sufficient to resolve it. The task hardware is a cabinet door whose hinge mechanism can be reconfigured by the experimenter to open in either the push or pull direction. The reconfiguration is internal to the hinge and is not visible from the door's exterior, so the opening direction is a hidden parameter that the robot must infer through interaction.

\paragraph{\slideblock.}
The robot must pick up a block from a fixed starting location, place it down on an inclined sliding surface at an inferred position along the slope, and release it; the block then slides down from rest under gravity and must come to rest overlapping a fixed target line lower on the slope. A trial is successful if the block stops overlapping the line; failures occur when the block overshoots or stops short. A smart policy should use the outcome of the first release, specifically how far the block slid relative to the target, to estimate the effective friction and adjust the release position on subsequent attempts, moving the release point higher up the slope if the block stopped short and lower if it overshot. The task hardware consists of a 3D-printed cubic block and the inclined sliding surface with the target line marked on it. The block exhibits anisotropic friction: the coefficient differs across its six faces, and even within a single face, it varies with sliding direction. The block's effective friction coefficient under the current grasp and placement is a hidden parameter that the robot must infer through interaction, since the release position required to reach the target depends on how far the block will slide before coming to rest.

\paragraph{\balance.}
The robot must grasp a bar, position it above a narrow pivot, and release it so that the bar remains balanced. A trial is successful if the bar does not fall off the pivot. A smart policy should use the direction in which the bar tips on a failed attempt to localize the center of mass, since tipping toward one end indicates that end is heavier, and shift the placement along the bar accordingly on the next attempt, converging on the balance point through bisection-like updates. The task hardware is a hollow bar and a pivot fixed to the workspace. The experimenter places an internal weight inside the bar and fixes it with clay at a position that varies across bars, producing a center of mass that is not visible externally. The center of mass is the hidden parameter that the robot must infer through interaction, since whether the bar balances or topples depends on where it is placed relative to the unknown center of mass.

\paragraph{\pinball.}
The robot must pull the handle of a vertical spring-loaded launcher to an inferred depth and release it, propelling a ball upward to a target height. A trial is successful if the ball reaches a marker placed 22\, cm above its rest position. A smart policy should use the peak height reached on the first launch to identify which of the three balls is in play, since each ball produces a distinct height response to a given pull depth, and then select the pull depth calibrated to that ball on subsequent attempts. The task hardware is a launcher consisting of a handle segment and a spring segment separated by a fixed plate, with the spring housed in the upper segment; pulling the handle downward draws the connecting rod through the plate and compresses the spring, and upon release, the spring drives the ball upward from its rest position. The experimenter selects a ball whose mass falls within the range 2\,g to 89\,g at the start of each trial, and the ball's mass is the hidden parameter the robot must infer. This range spans nearly two orders of magnitude, over which the mapping from ball mass to required pull depth is not monotonic: the lightest ball empirically requires the deepest pull, contrary to the naive expectation that heavier balls always require more force.

\begin{table*}[t]
\centering
\caption{\textbf{Real-world task setup.} Each task contains a hidden physical parameter that the robot must infer through interaction. Data is collected via tele-operation, reported as successes/failures.}
\label{tab:realworld_tasks}
\small
\begin{tabular}{lllc}
\toprule
Task & Hidden parameter & Robot & Trials (S/F) \\
\midrule
Open Door  & Push vs.\ pull direction       & WidowX (Mobile ALOHA) & 10/10 \\
Slide Block  & Block friction coefficient     & Piper-X                 & 5/20 \\
Balance    & Bar's center of mass           & Piper-X                 & 6/12 \\
Pinball    & Ball mass                      & Piper-X                 & 6/12 \\
\bottomrule
\end{tabular}
\end{table*}

\subsection{Inference hyperparameters.} 
\label{sec:apdx:real:hyperparam}
We follow the same hyperparameters used in the Meta-World System Identification Suite ($n=2$ candidate plans per replanning round). To mitigate a first-frame bias we observe under the low-data regime, where the generator produces nearly deterministic plans that fail to explore alternative hypotheses across replanning rounds, we additionally perturb the first-frame observation with pixel-level Gaussian noise $\epsilon \sim \mathcal{N}(0, \sigma^2 I)$ before passing it to the video generator. We use $\sigma=0.1$ for all real-world experiments; \Cref{tab:noise_ablation} ablates this choice on Pinball, where without noise ($\sigma=0.0$) the replanned video closely mirrors the failed attempt, while excessive noise ($\sigma=0.3$) corrupts the scene context. While this first-frame bias is an issue in our current case, we expect it to diminish when further scaled up. 

\begin{table}[t]
\centering
\caption{\textbf{Noise injection ablation on Pinball.} Success rates with different pixel-level Gaussian noise scales $\sigma$ applied to the first-frame observation at inference time.}
\label{tab:noise_ablation}
\begin{tabular}{lc}
\toprule
$\sigma$ & Success Rate \\
\midrule
$0.0$ & 4/10 \\
$0.1$ (default) & \textbf{7/10} \\
$0.3$ & 3/10 \\
\bottomrule
\end{tabular}
\end{table}

\subsection{Evaluation protocol.}
We evaluate the single-episode replanning success rate conditioned on an initial failure: for each trial, we initialize the environment with a randomly sampled hidden parameter, let the robot execute the initial plan from the video planner, and (provided the initial attempt fails) measure whether the next replanned attempt succeeds. We use this conditional metric rather than the main paper's ``number of replans until success'' because the real-world setup limits the number of trials we can collect per condition.

\section{Method Details}
\label{sec:apdx:method}

\subsection{Video plan generator and retrieval module}
\label{sec:apdx:method:planner}
To generate meaningful plans in environments with unknown dynamics, our system first hypothesizes the hidden parameters of the environment via the retrieval and refinement of a latent state embedding, which serves as a proxy for the current environment configuration. We first discuss our \textit{Video Plan Generator} and \textit{Retrieval Module}, which generate candidate plans based on observed images and state embeddings. The \textit{Video Plan Generator} predicts future frames conditioned on the current observation and, when available, the state embedding. For new objects without prior interactions, the generator operates solely on the observed image by conditioning on a learned ``null'' state embedding.

The {\it Retrieval Module} extracts state embeddings from prior interactions by encoding past videos and selecting a representative embedding for each configuration ID. At inference time, it encodes the current interaction video, computes distances to past embeddings from the dataset $\mathcal{D}$, and uses a softmax function to sample the most relevant state embedding. In addition, we use an additional identification model to continuously refine the selected state embedding to best fit current interaction videos.

\subsubsection{State embedding}
\label{sec:apdx:method:planner:object_embed}
To represent the hidden system parameters, we encode interaction videos from previous action executions with the same object using a context encoder $E$. The video embedding for each video $v_j$ is obtained as $e^v_j = E(v_j)$. In the dataset $\mathcal{D}$, multiple data tuples may share the same configuration ID $o_j$. Naively encoding each video independently would result in multiple, different embeddings for the same object. To address this, we preprocess the dataset by grouping all entries with the same configuration ID and selecting the encoding from one of the successful executions (where $s_j = 1$) to represent the object, resulting in a state embedding $e^o_j$.

\subsubsection{Video plan generator}
\label{sec:apdx:method:planner:video_generator}
The Video Plan Generator is used to generate a video containing $T$ future frames conditioned on the current observation and a state embedding. We follow the implementation of AVDC’s video planner, which is a first-frame conditioned video diffusion model utilizing a 3D U-Net architecture. The original implementation conditions on task features were obtained from encoding natural language task descriptions with text encoders such as CLIP-Text, rather than state embeddings. To adapt the method to our needs, we project the state embedding into the hidden dimension of the language features and treat the projected embedding as an additional language token. The parameters of the projection layer are trained end-to-end along with the video planner, which is optimized using a standard denoising diffusion loss. For all experiments in this paper, we choose the resolution of generated frames to be 128x128 and $T=7$. Each time we invoke the Video Plan Generator, $T=7$ future frames are generated, we then concatenate them with the first frame, resulting in an $T+1=8$ frames video plan.

\label{sec:apdx:video-model}
\begin{table}[t]
\centering
\caption{\textbf{Hyperparameters.} Comparison of configuration parameters for the Meta-World benchmark.}
\small
\setlength{\tabcolsep}{16pt}
\begin{tabular}{lccc}
\toprule
\textbf{num\_parameters} & 125M \\
\textbf{diffusion\_resolution} & (32, 32) \\
\textbf{target\_resolution} & (128, 128) \\
\textbf{base\_channels} & 128 \\
\textbf{num\_res\_block} & 2 \\
\textbf{attention\_resolutions} & (2, 4, 8) \\
\textbf{channel\_mult} & (1, 2, 3, 4) \\
\textbf{batch\_size} & 128 \\
\textbf{training\_timesteps} & 12k \\
\textbf{denoising\_timestep} & 100 \\
\textbf{sampling\_timestep} & 25 \\
\bottomrule
\end{tabular}

\label{tab:config_comparison}
\end{table}

Our video diffusion model adopts a U-Net architecture following \citep{ko2024learning}. The original implementation conditions on the first frame and the task name tokens, where the task tokens are encoded using CLIP-Text, to predict future frames. Building on this framework, we additionally condition the video generation process on past interaction videos. Specifically, we encode the past interaction video as described in \Cref{sec:apdx:method:planner:context_encoder}, and concatenate the resulting embedding with the encoded task tokens after projecting it to match the CLIP-Text output dimensionality (512 channels) using a trainable linear layer.

To improve computational efficiency, the video diffusion model generates a low-resolution video plan through the denoising process. This low-resolution plan is then upsampled to the target resolution using a separate super-resolution model that predicts residuals over the upsampled frames. The diffusion model is trained using the v-parameterization approach with a learning rate of 1e-4. Key hyperparameters for the video diffusion model are summarized in \Cref{tab:config_comparison}.

\subsubsection{Retrieval module}
\label{sec:apdx:method:planner:retrieval_module}
During inference, the Video Plan Generator needs to take in a state embedding extracted from prior interactions. To ensure that the method is general across different encoders $E$, we first normalize and align the dimensionality of the feature space constructed by video features $e^v_j$ by applying PCA to all extracted video features $e^v_j$, yielding a PCA projection $P$ and PCA-transformed features $e^{vp}_j = P(e^v_j)$ and $e^{op}_j = P(e^o_j)$.

Given the interaction video from a prior execution, $v^e$, we first encode it using the encoder $E$ and apply the PCA transformation, resulting in $e^{ep} = P(E(v^e))$. We then compute the distance between $e^{ep}$ and the video encodings from $\mathcal{D}$ as $logits_j = -dist_j = -K(e^{ep}, e^{vp}_j)$, where $K$ represents the distance function. For the choice of distance function used in retrieval $K_{ret}$, we used Cosine Distance for CLIP and DINO-based methods and L2 Distance for other methods. After calculating the logits, we compute the sampling probability of each output embedding by passing the logits through a softmax function: $p_j = \text{softmax}(logits_j / \tau)$, where $\tau$ is the temperature parameter. Finally, we sample the output state embedding $e = e^{op}_k$ according to the resulting probability distribution $k \sim \text{Categorical}(p_1, p_2, \dots, p_x)$. While such a retrieval-based state estimation formulation necessitates the data being available during test time, it enables training-free state estimation and guarantees that the Video Plan Generator never accepts out-of-distribution state embeddings during inference time.

\subsubsection{Refining state embedding}
\label{sec:apdx:method:planner:refine_obj_embed}

In addition to using retrieval to find a relevant state embedding for a video, we also use an optimization-based approach to find a refined state embedding corresponding to a specific interaction video.  Specifically, we introduce another generative model, referred to as the Identification model or the ID module, trained with the parameters shared with the Video Plan Generator, as shown in \Cref{fig:model}. The ID module is trained to generate all possible interaction outcomes, including both successful and unsuccessful ones. 

During replanning, we freeze the ID module parameters and optimize only the state embedding $e$ using a clean-video denoising loss. Let $v^{\mathrm{obs}}$ be the observed interaction video and $v_t=\alpha_t v^{\mathrm{obs}}+\sigma_t\epsilon$ its noisy version at diffusion timestep $t$, with $\epsilon\sim\mathcal{N}(0,I)$. Conditioned on $e$, the ID module predicts the clean interaction video $\hat v_0(v_t,t;e)$. The refinement loss for this noisy input is
\[
    \mathcal{L}_{\mathrm{refine}}(e;t,\epsilon)
    = \operatorname{MSE}\!\left(v^{\mathrm{obs}},\hat v_0(v_t,t;e)\right).
\]
Here, $\hat v_0$ is the clean-video prediction obtained by denoising the noisy observed interaction, rather than a video rollout sampled from independent noise. Gradients update only $e$ while the ID module weights remain fixed.

Since the ID module and the Video Plan Generator share the same embedding space for state embeddings, the optimized state embedding can be used directly to generate video plans. In the experimental section of the paper, we analyze the effect of this refinement procedure for state estimation when 1) this approach optimizes state embedding from randomly initialized vectors drawn from a unit variance Gaussian distribution, 2) the state embedding is set to the output of the Retrieval Module, and 3) these two approaches are combined, where this approach optimizes the output of the Retrieval Module.

\subsubsection{Latent feature space}
\label{sec:apdx:method:planner:context_encoder}

Given the experience dataset $\mathcal{D}$, the context encoder $E$ decides the feature space that the Retrieval Module and the Video Plan Generator operate on. To implement $E$, we experiment with taking pre-trained vision encoders that are known to give good semantic representations, such as CLIP \citep{radford2021learning} and DINOv2 \citep{oquab2024dinov}. To encode videos with these image encoders, we used a simple strategy that first encodes each frame independently and then concatenates the per-frame features, resulting in a single-vector video feature. A visualization of the constructed feature space is shown in \Cref{sec:apdx:feat_space}.

\subsection{Rejection module}
\label{sec:apdx:method:rejection_module}
While the modules introduced in \Cref{sec:apdx:method:planner} alone are sufficient to perform system identification essential for solving the tasks given a set of videos of interactions in the environment, the identified system parameters are often not perfect. This is especially the case when we only have a limited number of interactions with the environment. Thus, to effectively estimate the precise state of the environment, it is important that the system repeatedly actively interacts with the environment.
To actively interact with the environment, we use a video plan generator to generate execution plans. To encourage the planner to explore novel plans and avoid being overly reliant on suboptimal beliefs, we employed a simple rejection-based sampling method. In the Rejection Module, the previously failed plans are stored in the failed plan buffer $B_P$. At each planning round, we generate $N$ plans $( v^p_1, v^p_2, \dots, v^p_N )$ instead of one by conditioning the Video Plan Generator on $N$ independently sampled state embeddings from the Retrieval Module, yielding $N$ different plans with potentially different beliefs on environment parameter $\theta$.
We first calculate the distance to the nearest failed plan for each candidate $v^p_i$:
\[
    d(v^p_i, B_P) = \min_{F \in B_P} K(v^p_i, F)
\]
where \( K \) is the distance function. Empirically, we found that L2 distance on raw pixel space works surprisingly well already.
We then select the plan \( v^{p*} \) whose distance to its nearest failed plan is the largest (out of a set of generated video plans):
\[
v^{p*} = \arg\max_{v^p_i} d(v^p_i, B_P)
\]
The selected $v^{p*}$ is later converted to low-level control signals through the Action Module.
\Cref{sec:apdx:method:planner} and \Cref{sec:apdx:method:rejection_module} together form the RELIC replanning loop. The rejection module fills the interaction buffer with videos of failed attempts, and the retrieval module turns that buffer into a latent embedding that conditions the next plan.

\subsection{Action module}
\label{sec:apdx:method:action_module}

Given a generated video plan, to convert the video into continuous actions to execute, we follow the concept of dense object-tracking from AVDC to recover actions from the video plans. On certain interaction tasks, the object of interest is not applicable. For example, in a bar-picking task where the robot is required to grasp around the center of mass to succeed, knowing the future trajectory of the bar is not sufficient to decide the grasp location. In these tasks, we track the robot's wrist instead. After obtaining the object/robot wrist trajectory from the tracking results, we convert the trajectory into actions using hand-crafted heuristic policies. 

\subsection{Data collection process}
In simulation, we use scripted policies to collect offline interaction data. Specifically, the policy takes as input both the current observation and a “guessed” system parameter. For example, in the Bar Picking task, the guess is a one-dimensional scalar representing the position of the center of mass relative to the bar. If this guess matches the true underlying parameter of the environment (\eg, both are centered), the interaction results in a successful video; otherwise, it produces a failed interaction. These outcomes correspond directly to the success signal $s_i$ described in \Cref{sec:method}.

For each object, we perform multiple interactions with different guesses, generating several failed trials, followed by an interaction using the correct guess, which produces a successful trial. We then assign all trials from the same object the same configuration ID $o_i$ and terminate the interaction loop for this object. The policy then starts interacting with a new object by randomizing the initial configuration and hidden parameters. As for the real-world experiment, we simply used tele-operation to replace the guess-augmented scripted policy.

\section{Additional Experiment Results}
\label{sec:apdx:exp}

\subsection{Retrieval from Multiple Failed Interactions}
\label{sec:apdx:multiple_interaction_retrieval}

To examine whether earlier failures provide useful evidence for subsequent
retrieval, we compare using only the latest failed interaction (1 video) with using the full Failed Interaction Buffer (at most $M=14$ videos). This comparison tests whether retrieval should rely on the most recent interaction outcome or aggregate evidence from the entire history of failed attempts.

In the latest-failure setting, retrieval follows the procedure described in \Cref{sec:method:planner} using only the most recent failed interaction. In the
full-history setting, we apply the same scoring procedure to each failed interaction in the buffer and average the resulting scores before the softmax operation. All other components and evaluation settings remain unchanged.

\begin{table*}[t]
    \centering
    \caption{\textbf{Latest failure versus full interaction history.}
    We report the success rate (SR), the average number of replanning trials
    (Trials, mean $\pm$ SEM), and fail count (Fails) over 400 trials per
    setting, with \texttt{refine\_steps}$=100$ and
    \texttt{M}$=14$. The best result for each metric within each
    task is shown in bold.}
    \label{tab:interaction_history_retrieval}

    \small
    \setlength{\tabcolsep}{4pt}
    \vspace{3pt}

    \resizebox{\linewidth}{!}{
    \begin{tabular}{lccccccccc}
    \toprule
    & \multicolumn{3}{c}{\textbf{\pushbar}}
    & \multicolumn{3}{c}{\textbf{\pickbar}}
    & \multicolumn{3}{c}{\textbf{\slidebrick}} \\
    \cmidrule(lr){2-4} \cmidrule(lr){5-7} \cmidrule(lr){8-10}
    \textbf{Method}
    & SR (\%)$\uparrow$ & Trials$\downarrow$ & Fails$\downarrow$
    & SR (\%)$\uparrow$ & Trials$\downarrow$ & Fails$\downarrow$
    & SR (\%)$\uparrow$ & Trials$\downarrow$ & Fails$\downarrow$ \\
    \midrule
    Latest failure only(Ours)
    & 95.8 & 4.28 $\pm$ 0.20 & 17
    & 98.0 & \textbf{3.46} $\pm$ 0.17 & 8
    & \textbf{82.8} & \textbf{6.64} $\pm$ 0.54 & \textbf{69} \\

    Full failure history
    & \textbf{98.0} & \textbf{4.08} $\pm$ 0.19 & \textbf{8}
    & \textbf{99.0} & \textbf{3.46} $\pm$ 0.17 & \textbf{4}
    & 77.2 & 7.39 $\pm$ 0.51 & 91 \\
    \bottomrule
    \end{tabular}
    }

\end{table*}

Table~\ref{tab:interaction_history_retrieval} shows that aggregating across the full history improves performance on \pushbar and \pickbar, while on \slidebrick, using only the latest failed interaction is better. We hypothesize that this difference stems from retrieval reliability. \slidebrick's task-relevant information is concentrated in a few keyframes (\eg how much the block is pushed upwards) that our frame-subsampling encoder may miss, so two interactions with the same hidden parameter (\ie friction) can yield very different embeddings, making retrieval less accurate in the first few replans. Averaging the full history would then let these early errors bias every later round, whereas retrieving from only the latest failure resamples from the most recent outcome alone and avoids carrying them forward. These results suggest that history-based retrieval can help, but that its effectiveness depends on the quality of the underlying features. Therefore, stronger encoders or subsampling strategies for history-aware retrieval are an interesting direction for future work.

\subsection{Hidden State Optimization Performance}
\label{sec:apdx:hidden_state_optimization}

We evaluate how the prediction accuracy changes over successive replanning trials as additional interaction outcomes become available. For \pickbar and \pushbar, we measure the error in the predicted center-of-mass location. For \slidebrick, we measure the error between the predicted push height and the corresponding ground-truth target height.

\begin{figure}[htb]
    \centering
    \includegraphics[width=\linewidth]{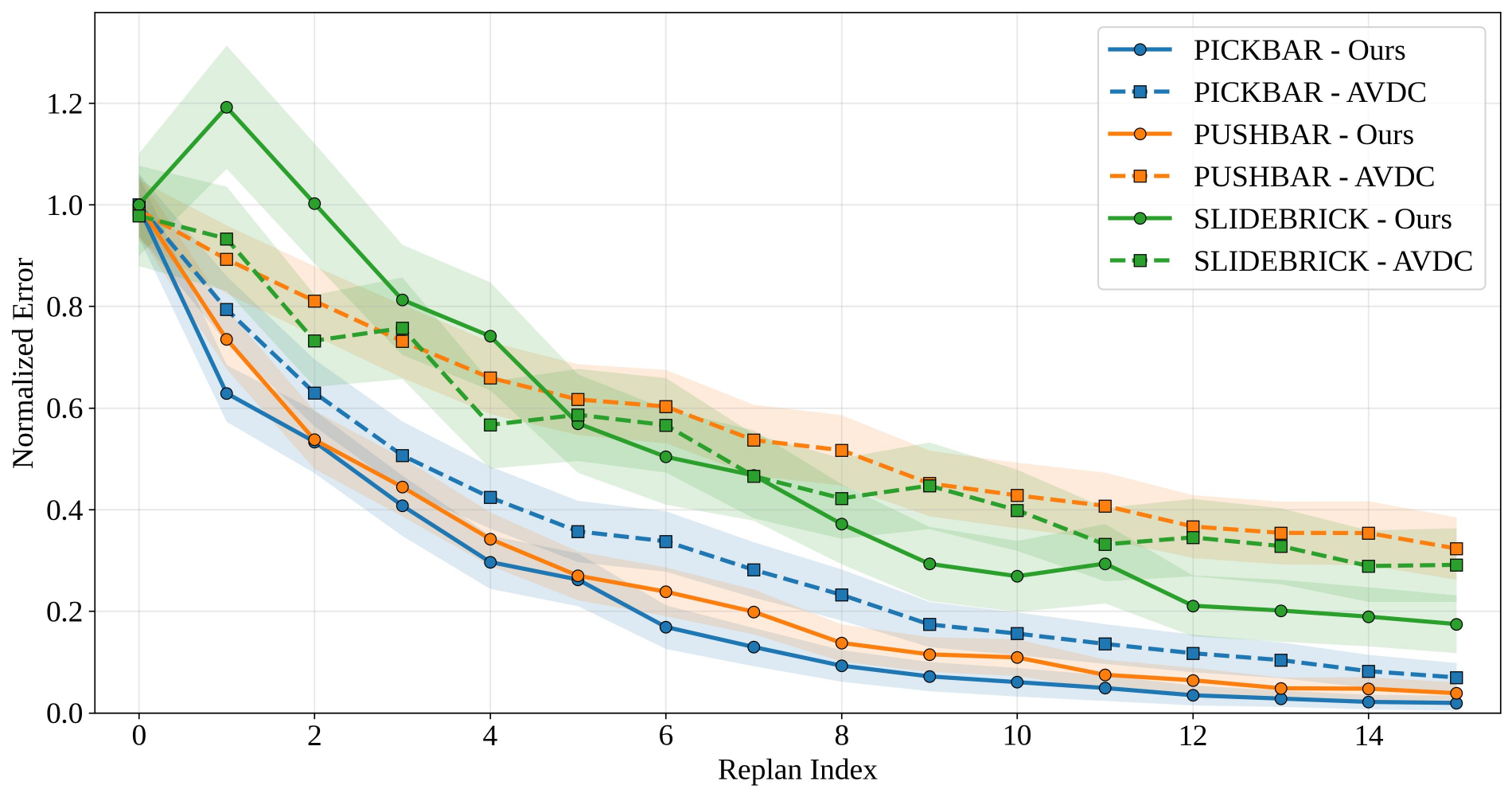}
    \caption{\textbf{Hidden-state optimization performance over replanning trials.}
    Normalized prediction error over replanning trials. Error is measured
    on the center-of-mass location for \pickbar and \pushbar, and on the target
    push height for \slidebrick. Shaded regions denote 95\% confidence
    intervals over 400 trials. Lower is better.
    }
    \label{fig:factor_error}
\end{figure}

Figure~\ref{fig:factor_error} shows that prediction error generally
decreases over successive replanning trials. Although our method does not always outperform the baseline in the early replans, the optimized state embedding is progressively refined as additional interaction outcomes are incorporated. As a result, our method achieves lower final prediction error on \pickbar, \pushbar, and \slidebrick. These results indicate that repeated interaction provides useful task-specific evidence for improving hidden-state estimation over time. 

\subsection{State Embedding Feature Space}
\label{sec:apdx:feat_space}
\begin{figure*}[t]
  \centering
  \begin{subfigure}[t]{0.32\textwidth}
    \centering
    \includegraphics[width=\linewidth]{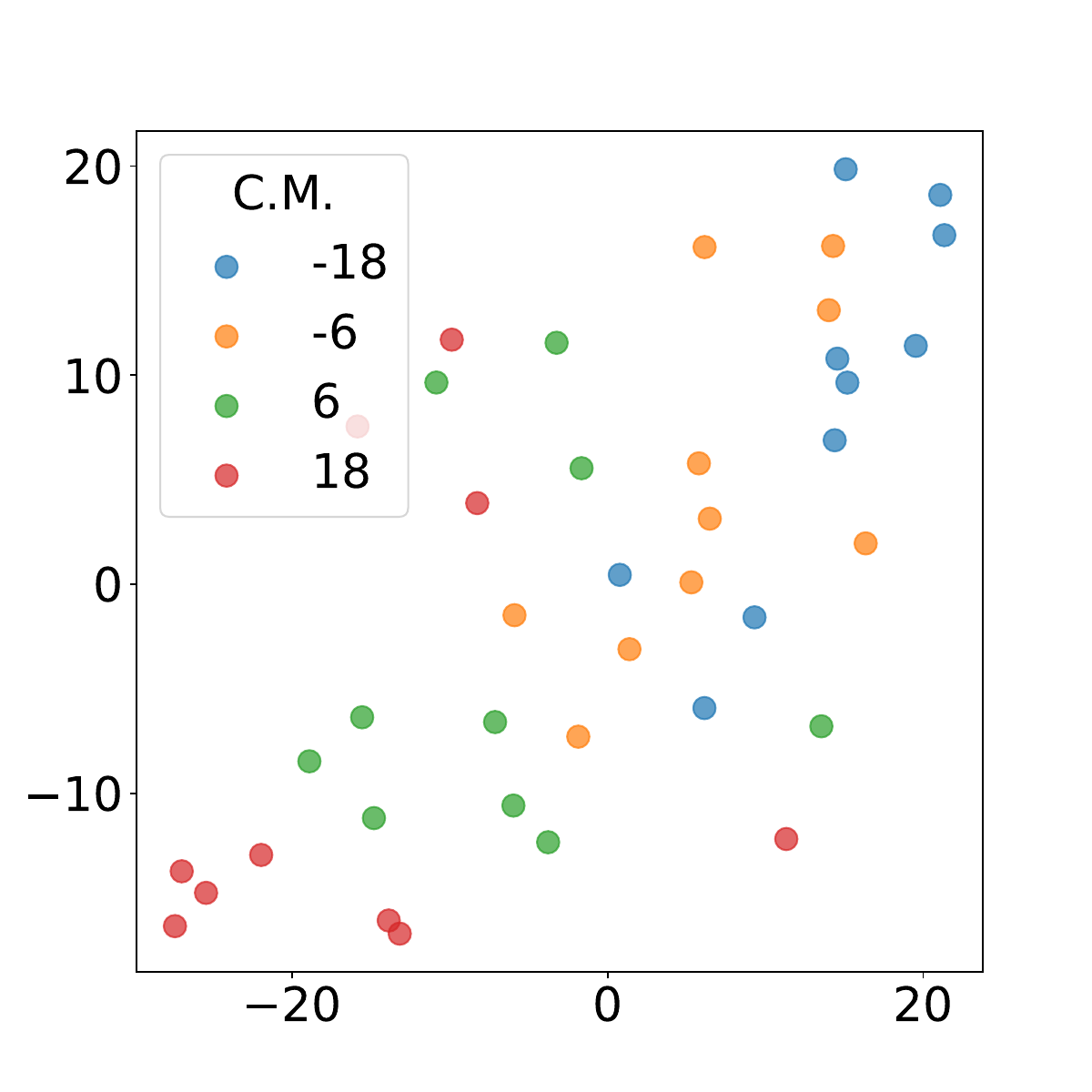}
    \caption{\pickbar (Continuous)}
  \end{subfigure}
  \hfill
  \begin{subfigure}[t]{0.32\textwidth}
    \centering
    \includegraphics[width=\linewidth]{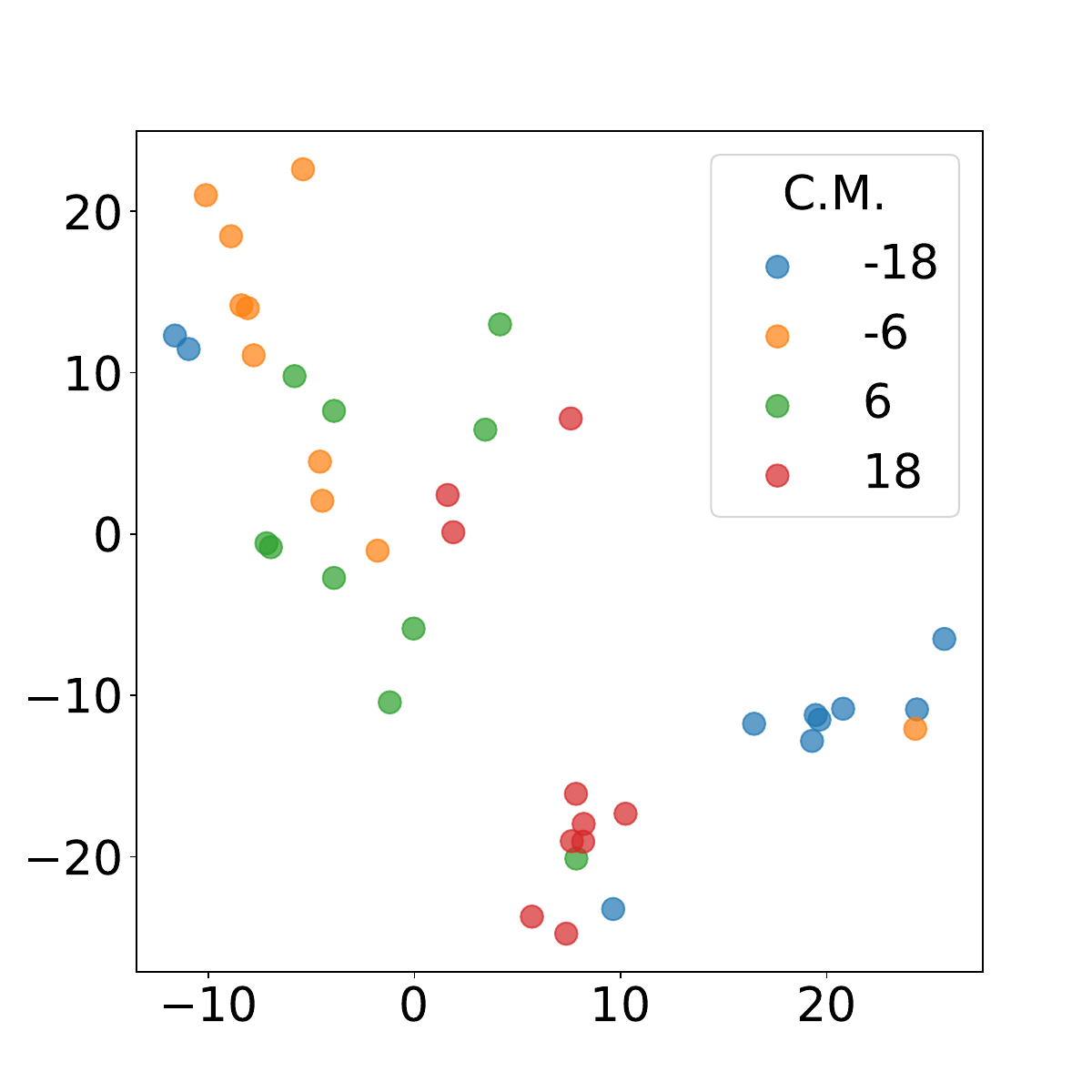}
    \caption{\pushbar (Continuous)}
  \end{subfigure}
  \hfill
  \begin{subfigure}[t]{0.32\textwidth}
    \centering
    \includegraphics[width=\linewidth]{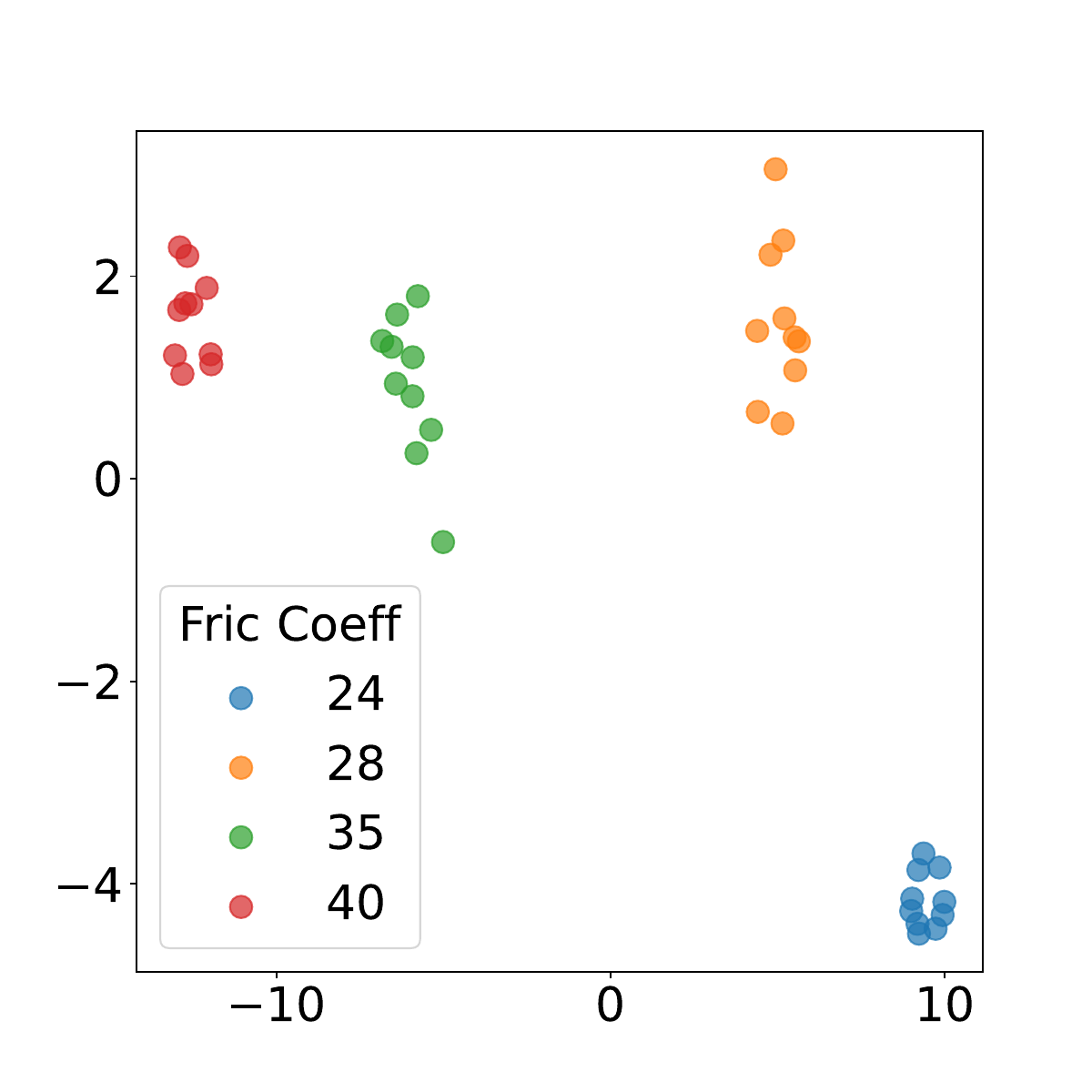}
    \caption{\slidebrick (Continuous)}
  \end{subfigure}

  \vspace{0.5em}

  \begin{subfigure}[t]{0.32\textwidth}
    \centering
    \includegraphics[width=\linewidth]{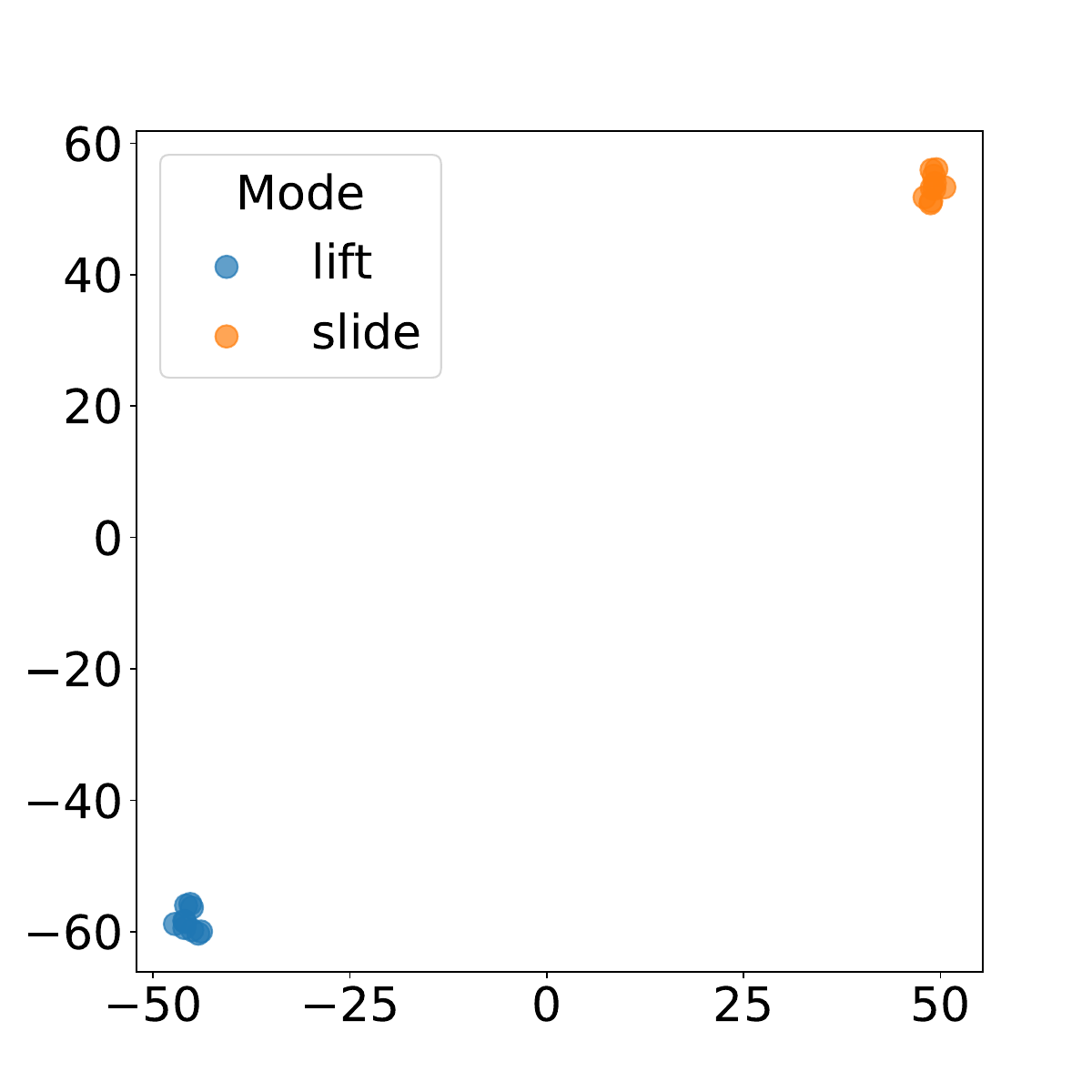}
    \caption{\openbox (Discrete)}
  \end{subfigure}
  \begin{subfigure}[t]{0.32\textwidth}
    \centering
    \includegraphics[width=\linewidth]{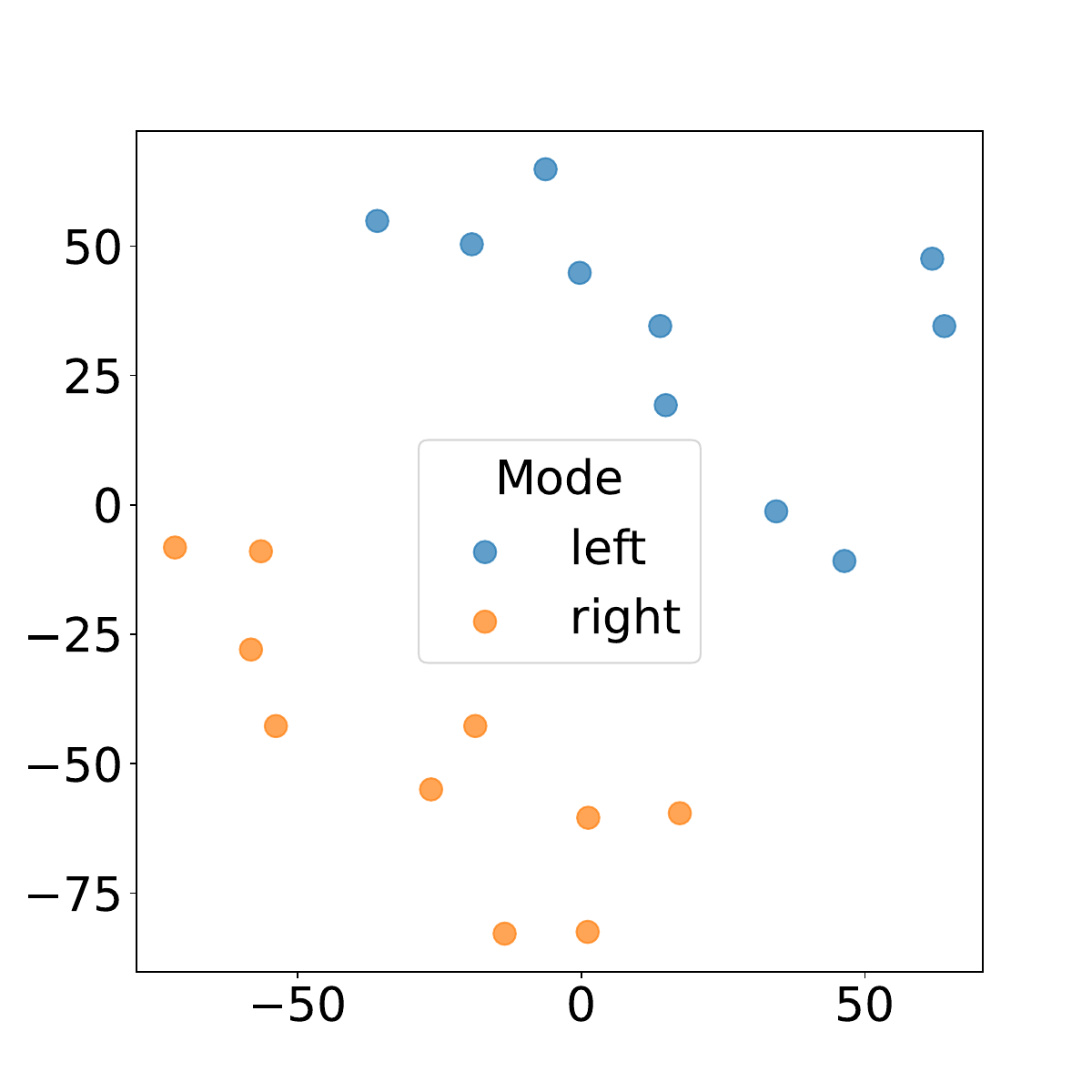}
    \caption{\turnfaucet (Discrete)}
  \end{subfigure}

  \caption{\textbf{Feature space visualization.} Top row: continuous parameter environments (\pickbar, \pushbar, \slidebrick). Bottom row: discrete parameter environments (\openbox, \turnfaucet).}
  \label{fig:feat_space}
\end{figure*}

\Cref{fig:feat_space} visualizes the feature space introduced in \Cref{sec:method:planner}, using DINOv2 \citep{oquab2024dinov} as the encoder. The plots are generated by applying t-SNE to sets of state embeddings from each environment, where for continuous modes we uniformly sample four values for visualization. Our method yields a well-structured feature space in which different modes can be separated without additional data processing, facilitating more effective downstream learning.

\subsection{Runtime analysis}
\label{sec:apdx:runtime-analysis}

\begin{table}[t]
\centering
\caption{\textbf{Computation time (in seconds)} averaged over 100 trials.}
\label{tab:runtime}
\begin{tabular}{lcc}
\toprule
Component & AVDC & Ours \\
\midrule
Retrieve & N/A & 0.062 $\pm$ 0.001 \\
Refine & N/A & 2.713 $\pm$ 0.010 \\
DDIM & 0.542 $\pm$ 0.001 & 0.546 $\pm$ 0.002 \\
Reject & N/A & 0.078 $\pm$ 0.001 \\
Optical Flow & 0.097 $\pm$ 0.000 & 0.097 $\pm$ 0.000 \\
\midrule
Total & 0.639 $\pm$ 0.001 & 3.496 $\pm$ 0.010 \\
\bottomrule
\end{tabular}
\end{table}

As shown in \Cref{tab:runtime}, we measured the average computation time over 100 iterations on a node with an Intel Xeon Platinum 8480 CPU and an H100 GPU. Each iteration includes embedding retrieval, video generation (DDIM), plan selection (reject), and action decoding via optical flow. \ICLRRevision{Although our full pipeline is slower per replan, this does not imply worse time-to-success. A more relevant metric for practical deployment is wall-clock time until the first successful execution, rather than latency per planning step. For example, in the \pushbar task, AVDC requires an average of 6.83 replans before success, with approximately 0.64 seconds per replan. Our full method requires fewer replans (4.26 on average), with approximately 3.50 seconds per replan.}
In addition, when a plan fails, execution requires approximately 25 seconds to terminate before replanning. Taking this into account, the expected time to first success is:
\begin{itemize}
    \item AVDC: 6.83 $\times$ (0.64 + 25) $\approx$ 175.1 seconds
    \item Ours: 4.26 $\times$ (3.50 + 25) $\approx$ 121.4 seconds
\end{itemize}
Our method reaches the first success substantially faster overall due to the lower failure rate, even accounting for the higher per-replan cost.
Moreover, our computational cost is amortizable, while execution time is not. With better hardware or a more accelerated algorithm, the refinement overhead can be reduced, whereas the real-world cost of a failed execution remains fixed. This means the time advantage of our method naturally increases with improved computing infrastructure.
Finally, our method provides an additional benefit that vanilla video planning does not: once a state embedding is identified for a particular object, it can be reused in future interactions. This allows the agent to succeed with higher probability on subsequent trials without repeating the identification process, whereas AVDC would again incur the full failure cost of approximately 175.1 seconds to reach its next success.

\subsection{Training Data and Scaling Effects.}
\label{sec:apdx:training_data}

The training data used in our main experiment is summarized in \Cref{tab:num_data}, where the "100\%" row indicates the amount of data per environment. To account for the scarcity of data in real-world settings and to assess the effect of data scale on performance, we repeated the experiment with only 28\% of the data. Results are presented in \Cref{tab:scaling_effect} and \Cref{fig:scaling_effect}.

\begin{table*}[t]
\centering
\caption{\textbf{Number of demonstrations used for each task.} 
Each cell shows failed/successful demonstrations.}
\label{tab:demo_counts}
\scalebox{0.9}{\begin{tabular}{lcccccc}
\toprule
 & \pushbar & \pickbar & \slidebrick & \openbox & \turnfaucet & Total \\
\midrule
Main exp (100\%) & 6000/250 & 6000/250 & 1560/130 & 20/20 & 20/20 & 13620/650 \\
28\% & 1000/250 & 1000/250 & 1300/130 & 20/20 & 20/20 & 3360/650 \\
\bottomrule
\label{tab:num_data}
\vspace{-0.5cm}
\end{tabular}}
\end{table*}

\begin{table*}[htb]
\centering
\caption{\textbf{Scaling effect on performance.} 
Reported metric is \emph{Replans until Success} 
($\downarrow$, lower is better). 
We compare performance with 28\% vs. 100\% data across each environment.}
\label{tab:scaling_effect}

\begin{subtable}{0.32\textwidth}
\centering
\caption{\pushbar}
\begin{tabular}{lcc}
\toprule
Method & 28\% & 100\% \\
\midrule
CQL   & 12.69 & 9.96 \\
BC    & 11.24 & 10.26 \\
AVDC  & 5.65  & 6.83 \\
Ours  & \textbf{3.91} & \textbf{4.26} \\
\bottomrule
\end{tabular}
\end{subtable}
\hfill
\begin{subtable}{0.32\textwidth}
\centering
\caption{\pickbar}
\begin{tabular}{lcc}
\toprule
Method & 28\% & 100\% \\
\midrule
CQL   & 11.57 & 9.22 \\
BC    & 8.88  & 8.79 \\
AVDC  & 3.60  & 4.41 \\
Ours  & \textbf{3.56} & \textbf{3.84} \\
\bottomrule
\end{tabular}
\end{subtable}
\hfill
\begin{subtable}{0.32\textwidth}
\centering
\caption{\slidebrick}
\begin{tabular}{lcc}
\toprule
Method & 28\% & 100\% \\
\midrule
CQL   & 12.33 & 8.94 \\
BC    & \textbf{7.34}  & 9.24 \\
AVDC  & 10.75 & 7.36 \\
Ours  & 11.42 & \textbf{6.69} \\
\bottomrule
\end{tabular}
\end{subtable}

\vspace{0.5em}

\begin{subtable}{0.32\textwidth}
\centering
\caption{\openbox}
\begin{tabular}{lcc}
\toprule
Method & 28\% & 100\% \\
\midrule
CQL   & 3.70 & 3.32 \\
BC    & 3.55 & 1.59 \\
AVDC  & 1.84 & 1.82 \\
Ours  & \textbf{1.39} & \textbf{1.25} \\
\bottomrule
\end{tabular}
\end{subtable}
\hfill
\begin{subtable}{0.32\textwidth}
\centering
\caption{\turnfaucet}
\begin{tabular}{lcc}
\toprule
Method & 28\% & 100\% \\
\midrule
CQL   & 4.68 & 5.63 \\
BC    & 4.15 & 3.54 \\
AVDC  & 1.78 & 1.67 \\
Ours  & \textbf{1.59} & \textbf{1.39} \\
\bottomrule
\end{tabular}
\end{subtable}
\hfill
\begin{subtable}{0.32\textwidth}
\centering
\caption{Overall (Normalized)}
\begin{tabular}{lcc}
\toprule
Method & 28\% & 100\% \\
\midrule
CQL   & 3.36 & 2.56 \\
BC    & 2.58 & 1.98 \\
AVDC  & 1.32 & 1.31 \\
Ours  & \textbf{1.16} & \textbf{1.00} \\
\bottomrule
\end{tabular}
\end{subtable}

\end{table*}

\begin{figure}[ht]
  \centering
  \begin{tikzpicture}
    \definecolor{tabblue}{HTML}{1f77b4}
    \definecolor{taborange}{HTML}{ff7f0e}
    \definecolor{tabgreen}{HTML}{2ca02c}
    \definecolor{tabred}{HTML}{d62728}
    \begin{axis}[
        width=0.6\linewidth,
        height=5.5cm,
        xtick={28,100},
        xticklabels={28\% Data, 100\% Data},
        grid=both,
        grid style={dashed,gray!30},
        legend style={at={(0.5,1.05)},anchor=south,legend columns=-1},
        ymin=0.8, ymax=3.6,
    ]
    \addplot+[mark=o, thick, color=tabblue, mark options={fill=tabblue, draw=tabblue}]
        coordinates {(28,3.36) (100,2.56)};
    \addlegendentry{CQL}
    \addplot+[mark=square*, thick, color=taborange, mark options={fill=taborange, draw=taborange}]
        coordinates {(28,2.58) (100,1.98)};
    \addlegendentry{BC}
    \addplot+[mark=triangle*, thick, color=tabgreen, mark options={fill=tabgreen, draw=tabgreen}]
        coordinates {(28,1.32) (100,1.31)};
    \addlegendentry{AVDC}
    \addplot+[mark=diamond*, thick, color=tabred, mark options={fill=tabred, draw=tabred}]
        coordinates {(28,1.16) (100,1.00)};
    \addlegendentry{Ours}
    \end{axis}
  \end{tikzpicture}
  \caption{\textbf{Overall scaling effect.} Reported metric is \emph{Replans until Success} ($\downarrow$ lower is better).}
  \label{fig:scaling_effect}
\end{figure}

With 28\% of the data, our method surpasses baselines in 4 out of 5 tasks and achieves the best overall performance by a substantial margin, and the aggregate results show a clear advantage. As more data becomes available, our approach improves further, scaling smoothly and consistently across tasks. Taken together, these results demonstrate that our method delivers consistent and robust gains, both under limited data and when scaling up.

\ICLRRevision{
To further evaluate robustness under incomplete mode coverage, we include a harder triple-faucet task, where the robot must sequentially turn 3 faucets without knowing which side to push from for each of the faucets. This creates a task with \textbf{eight distinct task/system modes} and a longer planning horizon. We vary the size of the dataset and evaluate whether our retrieval process can still recover the correct configuration.
}

\ICLRRevision{
The results in \Cref{tab:3-faucet} show that our method already provides meaningful improvements over the random baseline even with very small (not even fully covering the 8 modes of the task) datasets, and its performance scales steadily as more experience data becomes available. These findings indicate that our encoding method learns embeddings that remain robust in challenging, long-horizon environments, enabling the Retrieval Module to recover system parameters that closely match the true environment configuration.}

\begin{table*}[t]
\caption{\textbf{Data scaling effect on the 3-faucet task.} The table shows the exact match success rate between the hidden parameter of the current environment and that of the retrieved object embedding.} 
\begin{center}
\begin{tabular}{lcccc}
\toprule
Method $\backslash$ Dataset Size \\(Successful, Failed)
& (2, 14) & (10, 70) & (40, 280) & (120, 840) \\
\midrule
Random & 11.87 $\pm$ 1.28 & 11.87 $\pm$ 1.28 & 11.87 $\pm$ 1.28 & 11.87 $\pm$ 1.28 \\
\textbf{Ours (Retrieval)} 
& \textbf{18.59 $\pm$ 1.54} & \textbf{30.94 $\pm$ 1.83}
& \textbf{53.12 $\pm$ 1.97} & \textbf{67.66 $\pm$ 1.85} \\
\bottomrule
\end{tabular}
\end{center}
\label{tab:3-faucet}
\end{table*}

\subsection{Multi-Faucet Environment and Robustness of State Embedding}

\begin{figure}[t]
    \centering
    \includegraphics[width=0.3\textwidth]{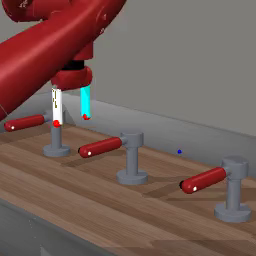}
    \caption{
    \textbf{Multi-faucet system identification setup.}
    A robot interacts with one of multiple faucet instances that share visual structure but differ in hidden physical properties (e.g., joint stiffness, friction, or rotation direction). 
    The agent must infer the correct interaction mode through probing actions rather than relying on appearance alone.
    }
    \label{fig:multi_faucet}
\end{figure}

\ICLRRevision{
To evaluate robustness in more complex and longer-horizon environments, we introduce an extended version of the faucet task in which the robot is required to turn three faucets sequentially. Similar to the original formulation, each faucet can only be rotated when pushed from the correct side; pushing from the wrong side causes the handle to jam and prevents further progress. This design induces $2^3 = 8$ distinct interaction modes and significantly increases task difficulty due to both extended horizon and compounding failure modes.}

\ICLRRevision{
We use this environment to examine whether the proposed state embedding encoding method can still recover the correct hidden system parameters under more challenging conditions. State embeddings are constructed using datasets of varying sizes, after which we apply the same Retrieval procedure as described in \Cref{sec:method:planner} using newly collected interaction trials. We then compare the retrieved embedding with the ground-truth hidden configuration of the environment and report the success rate of exact matches.}

\ICLRRevision{The results in \Cref{tab:3-faucet} demonstrate that the proposed encoding method already yields meaningful improvements over random selection with very limited data and continues to improve steadily as more experience becomes available. This indicates that the learned embeddings remain informative under long-horizon dynamics and combinatorial interaction modes, enabling the Retrieval Module to recover system parameters that closely match the underlying environment configuration.}

\subsection{Alternative plan selection strategy}

\ICLRRevision{
We consider an alternative plan selection strategy that uses the best retrieved successful video as a positive signal to guide planning. In this variant, the video retrieved by the Retrieval Module is treated as an attractor and directly influences plan selection by favoring candidates that are closer to this favorable outcome. We refer to this strategy as \emph{Attract}.} 

\ICLRRevision{
We compare this approach against our default Reject strategy, which explicitly filters out failed candidates through rejection during sampling. Both strategies share the same planning architecture and state embedding mechanism; the only difference lies in how retrieved information is used during selection. While Attract biases sampling toward favorable outcomes, Reject instead eliminates implausible or inconsistent candidates. The methods are evaluated over 80 trials for each task.}

\begin{table*}[t]
    \centering
    \caption{\textbf{Attract vs. Reject strategies.}}
    \label{tab:attract_vs_reject}
    \resizebox{\linewidth}{!}{
    \begin{tabular}{lcccccc}
        \toprule
        Method & \pushbar & \pickbar & \slidebrick & \openbox & \turnfaucet & Overall (Normalized) \\
        \midrule
        Attract & $5.55 \pm 0.55$ & $4.16 \pm 0.45$ & $12.26 \pm 0.52$ & $1.31 \pm 0.26$ & $1.31 \pm 0.29$ & $1.07$ \\
        \textbf{Reject (Ours)} 
                & $5.95 \pm 0.55$ 
                & $4.35 \pm 0.47$ 
                & $10.60 \pm 0.58$ 
                & $1.13 \pm 0.25$ 
                & $1.14 \pm 0.27$ 
                & $1.00$ \\
        \bottomrule
    \end{tabular}
    }
\end{table*}

\ICLRRevision{
The results are shown in Table~\ref{tab:attract_vs_reject}. Although Attract performs competitively on several tasks, it exhibits significantly worse performance on \slidebrick, the most challenging task in our evaluation suite. This indicates that a purely attraction-based signal may be insufficient under difficult dynamics where distinguishing incorrect plans is more helpful than favoring a single guess of a successful example.}

\ICLRRevision{
Note that rejection and attraction are not mutually exclusive. Rejection provides a repulsive signal that removes incompatible hypotheses, while attraction offers a directional signal toward plausible regions of the solution space. A combined approach that integrates both signals may further improve sampling efficiency and accelerate adaptation, which we leave as an interesting direction for future work.}

\subsection{Robustness to Task-Irrelevant Visual Distractors}
\label{sec:apdx:rejection_metric}
\begin{figure*}[t]
    \centering
    \begin{subfigure}{0.48\textwidth}
        \centering
        \includegraphics[width=0.6\textwidth]{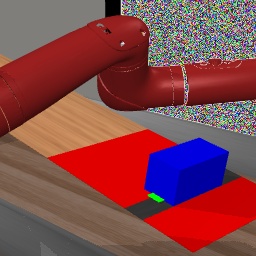}
        \caption{Pixel-level noise}
        \label{fig:noisy_env_orig}
    \end{subfigure}
    \hfill
    \begin{subfigure}{0.48\textwidth}
        \centering
        \includegraphics[width=0.6\textwidth]{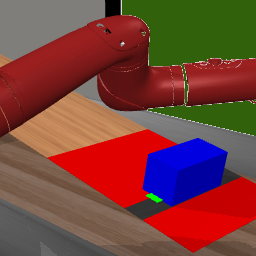}
        \caption{Uniform noise}
        \label{fig:noisy_env_rand}
    \end{subfigure}
    \caption{
    \textbf{Robustness to task-irrelevant visual noise.}
    We introduce appearance-randomized environments with visually distracting backgrounds while preserving identical interaction dynamics.
    This setting isolates whether distance functions respond to task-relevant motion rather than background variation.
    }
    \label{fig:randomized_env}
\end{figure*}

\ICLRRevision{
To evaluate robustness under visually complex conditions, we introduce two noisy variants of the \slidebrick and \openbox environments by randomly altering background appearances to simulate task-irrelevant visual signals. Such distractors are common in real-world settings and may interfere with similarity metrics that rely on raw pixel comparisons. Table~\ref{tab:noisy_env} reports the number of replans required before success under these noisy conditions.}

\begin{table}[t]
    \centering
    \caption{\textbf{Performance under visually noisy environments.} Metric: Replans until success.}
    \label{tab:noisy_env}
    \begin{tabular}{lcc}
        \toprule
        Method & \slidebrick Noisy & \openbox Noisy \\
        \midrule
        No Rejection   & $6.09$ & $1.91$ \\
        DINO           & $5.56$ & $\mathbf{1.65}$ \\
        L2 error       & $4.89$ & $1.79$ \\
        Seg + DINO     & $5.54$ & $\mathbf{1.65}$ \\
        Seg + L2 error & $\mathbf{4.53}$ & $\mathbf{1.65}$ \\
        \bottomrule
    \end{tabular}
\end{table}

\ICLRRevision{
The results in \Cref{tab:noisy_env} show that while L2 distance underperforms DINO on \openbox, it performs better on \slidebrick. A plausible explanation is that DINO embeddings prioritize semantic abstraction and may discard fine-grained spatial information that is critical for estimating hidden parameters in tasks that require high geometric precision. In contrast, pixel-level L2 distance preserves low-level structural cues that remain informative even under moderate visual noise.}

\ICLRRevision{
Furthermore, we demonstrate a simple and practical strategy for mitigating task-irrelevant signals by introducing segmentation to isolate task-relevant regions. By masking out background distractors, the distance metric, whether based on L2 or DINO features, can focus on the portions of the image that encode the underlying system parameters. This approach is feasible in practice, as high-quality segmentation masks can be obtained using modern vision-language models such as Segment Anything Models (SAMs). These results suggest that segmentation provides an effective mechanism for improving robustness in visually cluttered environments without requiring changes to the overall planning framework.}

\subsection{Generalization of state embeddings to Objects Variance}

\begin{figure*}[t]
    \centering
    \begin{subfigure}{0.48\textwidth}
        \centering
        \includegraphics[width=0.6\textwidth]{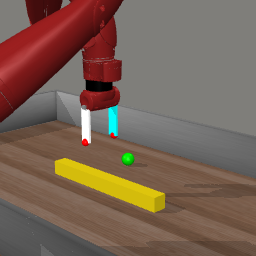}
        \caption{Appearance-randomized object 1}
        \label{fig:obj_orig}
    \end{subfigure}
    \hfill
    \begin{subfigure}{0.48\textwidth}
        \centering
        \includegraphics[width=0.6\textwidth]{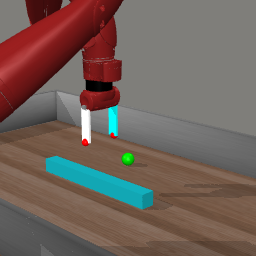}
        \caption{Appearance-randomized object 2}
        \label{fig:obj_rand}
    \end{subfigure}
    \caption{
    \textbf{Appearance variation under fixed interaction dynamics.}
    Object color and texture are randomized while geometry and physical properties remain unchanged.
    This setting decouples visual appearance from interaction behavior and tests adaptation under perceptual aliasing.
    }
    \label{fig:randomized_objects}
\end{figure*}

\ICLRRevision{
Our approach improves adaptation by retrieving state embeddings from previously observed objects and using them to approximate the hidden system parameters of the environment. When encountering a new object at test time, interaction mechanisms that were learned before (for example, push vs. pull for doors, or slide vs. lift for lids) can be recalled and transferred to guide behavior.}

\ICLRRevision{
To evaluate generalization, we construct novel environments in which object appearances differ from training, while the underlying interaction mechanisms remain unchanged. Each stored state embedding is associated with its ground-truth system parameters for evaluation. During testing, we retrieve the state embedding using the same retrieval procedure as in the main experiment and compare its associated parameters with those of the current environment. We consider the following baselines:}

\textbf{Random Guess:} Randomly select one embedding from the database.

\textbf{RELIC-Seen:} Perform retrieval using embeddings constructed from environments seen during training.

\textbf{RELIC-Unseen:} Perform retrieval on novel environments with unseen object appearances.

\begin{table*}[t]
    \centering
    \caption{Raw retrieval error on hidden system parameters (lower is better).}
    \label{tab:ise_raw}
    \scalebox{0.9}{\begin{tabular}{lccccc}
        \toprule
        Method & \pushbar & \openbox & \pickbar & \turnfaucet & \slidebrick \\
        \midrule
        Random Guess 
            & $.0954 \pm .0052$ 
            & $.4900 \pm .0500$ 
            & $.0954 \pm .0052$ 
            & $.4900 \pm .0500$ 
            & $.0528 \pm .0040$ \\
        RELIC-Seen
            & $.0694 \pm .0057$
            & $.0200 \pm .0140$ 
            & $.0793 \pm .0069$ 
            & $.0200 \pm .0140$ 
            & $.0448 \pm .0032$ \\
        RELIC-Unseen
            & $.0693 \pm .0049$
            & $.0700 \pm .0255$ 
            & $.0729 \pm .0054$ 
            & $.0200 \pm .0140$ 
            & $.0428 \pm .0032$ \\
        \bottomrule
    \end{tabular}}
\end{table*}

The results in \Cref{tab:ise_raw} show that most of the performance gains of our method can be generalized to unseen cases.

\section{Probabilistic intuition for rejection.}
A simple sampling argument illustrates why drawing multiple candidate plans helps. If the correct interaction mode has a belief probability $q_{k^\star}$, then drawing $N$ independent hypotheses contains the correct mode with probability $1 - (1 - q_{k^\star})^N$, strictly greater than $q_{k^\star}$ for any $N \geq 2$. Under the assumption that the rejection distance reliably distinguishes correct from incorrect modes. This amplifies the per-step success probability and yields exponential decay in the probability of failure over $M$ attempts. We emphasize that this is an intuition rather than a guarantee: the practical effectiveness, the degree of rejection, depends on the distance function $K_{rej}$ separating correct from incorrect rollouts, which we study in~\Cref{sec:exp:ablation} empirically.

\end{document}